\pdfoutput=1
\documentclass[11pt]{article}

\usepackage[preprint]{acl}
\usepackage{xcolor}
\usepackage{times}
\usepackage{latexsym}
\usepackage[T1]{fontenc}
\usepackage[utf8]{inputenc}
\usepackage{microtype}
\usepackage{inconsolata}
\usepackage{colortbl}
\usepackage{adjustbox}
\usepackage{diagbox}
\usepackage{amsmath,amssymb, amsthm}
\usepackage[inline]{enumitem}
\usepackage{array,multirow,graphicx,makecell}
\usepackage{booktabs}
\usepackage{tabularx}
\usepackage{xfrac}
\usepackage[skins]{tcolorbox}
\tcbuselibrary{breakable}
\usepackage{anyfontsize}
\usepackage[toc,page,header]{appendix}
\usepackage{minitoc}

\newtheorem{theorem}{Theorem}
\theoremstyle{remark}
\newtheorem{remark}{Remark}

\definecolor{sd_orange}{RGB}{250, 99, 38}
\definecolor{td_blue}{RGB}{0, 112, 192}
\definecolor{st_green}{RGB}{56, 140, 114}
\definecolor{tt_blue}{RGB}{31, 78, 121}
\definecolor{ss_orange}{RGB}{197, 90, 17}
\definecolor{mgreen}{RGB}{112, 173, 71}

\newcommand{\E}{\mathbb{E}}

\newcommand{\Var}{\mathrm{Var}}
\newcommand{\Cov}{\mathrm{Cov}}
\newcommand{\sss}{\textcolor{ss_orange}{\mathrm{SS}}}
\newcommand{\st}{\textcolor{st_green}{\mathrm{ST}}}
\newcommand{\ttt}{\textcolor{tt_blue}{\mathrm{TT}}}
\newcommand{\sd}{\textcolor{sd_orange}{\mathrm{SD}}}
\newcommand{\td}{\textcolor{td_blue}{\mathrm{TD}}}
\newcommand{\idd}{\mathrm{IDD}}
\newcommand{\drop}{\overline{\Delta}}
\newcommand{\triplet}{(\sss, \ttt, \st)}

\title{Measuring the Robustness of NLP Models to Domain Shifts}

\author{Nitay Calderon\Thanks{ Equal contribution. \\  \url{https://github.com/nitaytech/DomainRobustness}}, Naveh Porat\footnotemark[\value{footnote}], Eyal Ben-David, Alexander Chapanin, \\ \textbf{Zorik Gekhman, Nadav Oved, Vitaly Shalumov \and Roi Reichart} \\
  Technion - Israel Institute of Technology  \\
{\{nitay@campus.$|$roiri@\}technion.ac.il}
}

\begin{document}
\maketitle

\doparttoc 
\faketableofcontents 

\begin{abstract}
Existing research on Domain Robustness (DR) suffers from disparate setups, limited task variety, and scarce research on recent capabilities such as in-context learning. Furthermore, the common practice of measuring DR might not be fully accurate. Current research focuses on challenge sets and relies solely on the Source Drop (SD): Using the source in-domain performance as a reference point for degradation. 
However, we argue that the Target Drop (TD), which measures degradation from the target in-domain performance, should be used as a complementary point of view. 
To address these issues, we first curated a DR benchmark comprised of 7 diverse NLP tasks, which enabled us to measure both the SD and the TD. 
We then conducted a comprehensive large-scale DR study involving over 14,000 domain shifts across 21 fine-tuned models and few-shot LLMs.
We found that both model types suffer from drops upon domain shifts. While fine-tuned models excel in-domain, few-shot LLMs often surpass them cross-domain, showing better robustness.
In addition, we found that a large SD can often be explained by shifting to a harder domain rather than by a genuine DR challenge, and this highlights the importance of TD as a complementary metric.
We hope our study will shed light on the current DR state of NLP models and promote improved evaluation practices toward more robust models.
\end{abstract}

\section{Introduction}
\label{sec:intro}

\begin{figure}[ht!]
    \centering
    \includegraphics[width=0.48\textwidth]{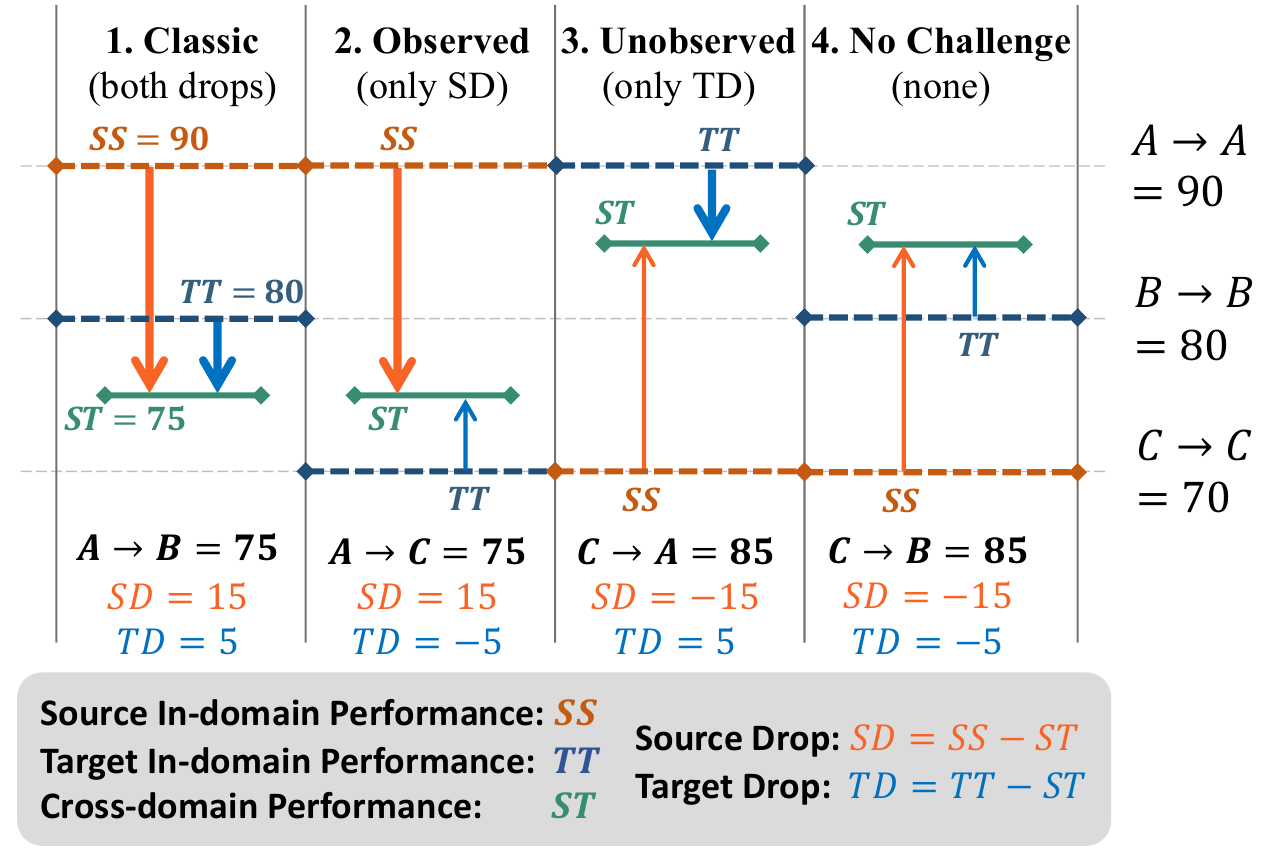}
    \caption{Illustration of the four domain shift scenarios. In the \textit{Classic} and \textit{Observed} scenarios, we observe a 15-point drop between the \textit{Source In-domain Performance} ($\sss$) and the \textit{Cross-domain Performance} ($\st$).
    Conversely, in the \textit{Unobserved} and \textit{No Challenge} scenarios, $\sss=70$ and $\st=85$, meaning the model gains 15 points upon domain shift.
    We would typically conclude that there is a DR challenge only in the first two scenarios. However, we argue that this commonly adopted perspective is inaccurate since it overlooks the \textit{Target In-domain Performance} ($\ttt$). Our work provides a fresh perspective by considering both degradation metrics: The \textit{Source Drop} ($\sd$) and the \textit{Target Drop} ($\td$). 
    }
    \label{fig:intro}
    \vspace{-0.7em}
\end{figure}

Modern transformer-based NLP models, and particularly \textit{Large Language Models (LLMs)} have proven effective on various tasks and evaluation setups, including fine-tuning \citep{bert, t5} and in-context learning \citep{gpt3, palm}. Following that, there has been an improvement in the models' ability to perform tasks while transferring to domains with no labeled data available \citep{transformers_improve_robustness, pada, zero_shot_arch}. Despite these improvements, the performance upon domain shift can still be inferior to the model's performance on the source domains, a problem we refer to as the \textit{Domain Robustness (DR) challenge} \citep{unsupervised_da_survey, robustness_nlp_survey, da_hu23, da_ya23}.

Research of DR is quite disparate: A wide variety of setups, models, training procedures, and different dataset sizes are used. There is also a severe lack of variety in evaluation tasks for DR: Most papers use classification tasks, omitting important tasks such as sequence tagging, question answering, and text generation \citep{transformers_improve_robustness, wilds}. Moreover, many past works use challenge sets to measure the DR challenge. These are highly curated datasets that select synthetic \citep{synthetic_nmt, wildnlp} or particularly hard samples for models to process under domain shifts \citep{nli_hans, revisiting}.
All this makes it hard to compare different works and map out the extent of the DR challenge in a \textit{natural domain shift setting}. 

Moreover, prior works focused solely on fine-tuned models, disregarding few-shot setups that have become prominent.\footnote{We use \textit{few-shot models} to denote LLMs in an in-context learning setting, where the prompt contains demonstrations.} 
In those setups, the DR challenge manifests itself more moderately: No training data can potentially anchor the model to the source distribution, but only a few demonstrations from the source domain are used in the prompt \citep{prompt_sensitivity, prompt_robustness}.

Adding to the above, we observe a fundamental problem with how we examine the DR challenge. Let us conduct a thought experiment, illustrated in Figure~\ref{fig:intro}: A model is trained and tested on data from domain A ($A \rightarrow A$), achieving a score of 90, but when tested on domains B and C, it scores 75. The observed 15-point drop typically leads to the conclusion the model lacks robustness, a common assertion in DR papers. But what if we were told that ``had the model been trained and tested on data from B, it would have achieved a score of 80'', would we still consider it as facing a severe DR challenge, given only a 5-point drop from B's in-domain performance ($B \rightarrow B$), rather than 15? Furthermore, if the model attains a score of 70 when trained and tested on domain C ($C \rightarrow C$), can we still assert a DR challenge exists even when it performs better cross-domain ($A \rightarrow C$)?

Building on the insights from the thought experiment, our paper introduces a novel perspective on the DR challenge. Traditional approaches typically focus on the \textcolor{sd_orange}{\textbf{Source Drop (SD)}}, assessing how model performance degrades compared to its source in-domain performance. However, this view overlooks the degradation compared to the setup where the model had been trained and tested on the target domain, which we define as the \textcolor{td_blue}{\textbf{Target Drop (TD)}}. We study these variables and build various metrics upon them in \S\ref{sec:methods}. 

Importantly, most works focus solely on the $\sd$ and overlook the $\td$, resulting in a partial depiction of the DR challenge. For instance, in studies involving challenge sets that report a large $\sd$, the drop may be primarily attributed to shifting to a harder domain ($\ttt < \sss$, see  \S\ref{sub:metrics}), and not by a genuine DR challenge, e.g., the \textit{Classic} and \textit{Observed} scenarios in Figure~\ref{fig:intro}.
By incorporating both metrics, we aim to provide a more holistic and accurate understanding of the DR challenge. 

To overcome deficiencies in the current body of research, in \S\ref{sec:benchmark} we introduce a novel DR benchmark. Unlike existing benchmarks, which largely rely on synthetic, adversarial, or challenge sets that may not adequately represent natural settings, our benchmark is unique and possesses four key properties: (i) It focuses on shifts (such as topical shifts) that naturally occur in real-life scenarios; (ii) It covers a wide variety of NLP tasks, more than other studies, including sequence and token level classification, QA, and generation tasks; (iii) Each task consists of several domains; and (iv) Each domain has a sufficient amount of labeled data, enabling its use as a source and as a target domain. 

Following that, we conduct an extensive study by benchmarking many fine-tuned models and few-shot LLMs, detailed in \S\ref{sec:experimental_setup}. We examine factors such as the model size, dataset size, number of few-shot demonstrations, and more. Our findings, reported in \S\ref{sec:results}, incorporate results of more than 14,000 domain shifts of 21 models and various training and testing setups. Our main findings are: 
\begin{enumerate}[itemsep=1pt,parsep=1pt, topsep=1pt, align=left,leftmargin=*]
\item Fine-tuned models suffer from drops upon domain shifts. While the extent of the drop varies, challenging shifts are prevalent in every task;
\item Increasing the size of fine-tuned models enhances both in-domain and cross-domain performance while reducing performance drops, particularly in classification tasks;
\item Few-shot models also face a DR challenge as the domain of the demonstrations impacts their performance. However, the domain shift effect for few-shot models is weaker and more nuanced;
\item Increasing the fine-tuning dataset size as well as the number of few-shot demonstrations enhances in-domain and cross-domain performance but can also mildly increase the drop due to stronger ``source domain anchoring'';
\item While fine-tuned models excel in-domain, few-shot LLMs often surpass them cross-domain, showing better robustness and smaller drops;
\item Considering only one metric can lead to wrong conclusions since many domain shifts are not \textit{Classic}, and only one drop metric ($\sd$ or $\td$) is positive while the other is negative;
\item We found that a large $\sd$ can often be explained by shifting to a harder domain, and not by a genuine DR challenge;
\item Our focus on many natural domain shifts reveals that while challenge sets are helpful diagnostic tools, they tend to overestimate the severity of DR, which is generally milder;
\end{enumerate}

In conclusion, we show that thoroughly assessing DR in NLP models requires evaluating multiple domain shifts and incorporating both drop metrics ($\sd$ and $\td$). 
We manifest that while nuanced, the DR challenge is still prevalent.
In \S\ref{sec:discussion}, we delve into the implications of our findings for the NLP community. In Appendix \S\ref{sec:theorem}, we present a theorem that elucidates some of our findings regarding the relationship between the DR metrics. We hope this work will provide a fresh perspective on model robustness and facilitate further research.

\section{Related Work}
\label{sec:related}

The term DR generally refers to the extent to which the performance of a model does not degrade when applied to newly collected samples from other domains. In some cases, robustness refers to consistency (low variance) \citep{measuring_robustness}. Literature on robustness in NLP can be categorized by the type of distribution shift examined: Synthetic and Natural \citep{robustness_nlp_survey, da_hu23}. 

\textit{Synthetic shift} works include adversarial attacks \citep{adversarial_bert}, input perturbations \citep{synthetic_nmt}, counterfactual \citep{counterfactual_set}, diagnostic \citep{superglue} and challenge (or contrast) sets \citep{nli_hans}. 
These works assess robustness using datasets designed to challenge NLP models rather than represent a natural language distribution. While the synthetic shifts are helpful diagnostic tools \citep{dr_gym}, they do not accurately depict the actual state of DR ``in the wild''. Hence, we focus on natural domain shifts. 

\textit{Natural shift} study focuses on organic scenarios where a discrepancy exists between the training and deployment data. These studies encompass various setups, including medium shift \citep{robustness_qa}, temporal shift \citep{robustness_reddit_popularity_prediction}, and domain shift (e.g., to medical \citep{dg_example_health_ner} and legal \citep{dg_example_legal} domains). 

Researchers proposed various benchmarks to evaluate the robustness of NLP models and the quality of solutions, including domain shifts in a single NLP task \citep{multi_woz, lm_da_benchmark, robustness_qa,  adaptsum_summarization_dataset, qmsum_meeting_summarization_dataset, lm_da_benchmark_2, MarCQAp, yu-etal-2023-alert}, with challenge sets \citep{wildnlp, ft_vs_fs, prompt_robustness, revisiting} or only with fine-tuned models \citep{transformers_improve_robustness, dr_tu, wilds}. Our study addresses a broad range of domain shifts in many more NLP tasks than previous work, including sequence and token-level classification, QA, and generation. In addition, we examine both small fine-tuned models and few-shot LLMs. Importantly, unlike other works, which focused on the source drop, we also consider the target drop, providing a more holistic perspective on DR. To the best of our knowledge, this is the most comprehensive DR study in NLP.

\section{Domain Robustness}
\label{sec:methods}


\textit{Domain} is a widely used term in NLP that typically refers to a cohesive corpus or dataset, which may be characterized by factors such as topic, style, genre, syntax, linguistic register, and medium. Although `domain' lacks a clear and consistent definition \citep{unsupervised_da_survey}, we formally describe a \textit{domain} $\mathcal{D}$ by a joint distribution $P_{\mathcal{D}}(X,Y)$ over $\mathcal{X}$ (the input space) and $\mathcal{Y}$ (the outcome space). In a \textit{domain shift}, the source domain $\mathcal{S}$, and the target domain $\mathcal{T}$ differ in their underlying joint distribution $P_{\mathcal{S}}(X,Y) \ne P_{\mathcal{T}}(X,Y)$. 

Given a training set of examples from the source domain $S \sim \mathcal{S}$, the goal of the NLP model is to learn $P_{\mathcal{S}}(X,Y)$ (or $P_{\mathcal{S}}(Y|X)$), and to the generalize to the (potentially unknown) target domain distribution(s) in which it will be deployed, $P_{\mathcal{T}}(X,Y)$. To evaluate the performance on the target domain, we use a test set $T \sim \mathcal{T}$, which is \textit{unobserved during training}. We use the term \textit{Domain Robustness (DR)} to describe \textbf{the inherent (in)ability of an NLP model to generalize from the source domain to the target domains}. 

For fine-tuned models, the DR challenge arises when the test data comes from a domain that is different from the labeled training data. Meanwhile, few-shot models face the DR challenge when the domain of the demonstrations used in the prompt differs from that of the target data. 

\begin{table}
\centering
\begin{adjustbox}{width=0.46\textwidth}
\begin{tabular}{m{0.06\textwidth}|p{0.43\textwidth}}
\toprule
$\sss$ & Source In-domain Performance \\
$\ttt$ & Target In-domain Performance \\
$\st$ & Cross-domain Performance \\
\hline
$\sd$ & Source Drop (Observed Drop): $\sss-\st$ \\
$\td$ & Target Drop (Unobserved Drop): $\ttt-\st$ \\
$\idd$ & In-domain difference: $\sss-\ttt$ \\
\hline
{\footnotesize $\overline{\sss}$} & Average In-domain: $\E[\sss] = \E[\ttt]$ \\
{\footnotesize $\overline{\st}$} & Average Cross-domain: $\E[\st]$ \\
{\footnotesize $\drop$} & Average Drop: $\overline{\sss} - \overline{\st} = \E[\sd] = \E[\td]$ \\
{\footnotesize $W_{\sd}$} & Worst $\sd$: $\max_{(S,T)}{\sd}$ \\
{\footnotesize $W_{\td}$} & Worst $\td$: $\max_{(S,T)}{\td}$ \\
\bottomrule
\end{tabular}
\end{adjustbox}
\caption{The notations of Domain Robustness concepts and metrics we use in this study. Toy example in Table~\ref{tab:toy}.}
\label{tab:notations}
\vspace{-0.5em}
\end{table}

\subsection{Measuring Domain Robustness}
\label{sub:metrics}

This subsection proposes concepts and metrics for characterizing the DR challenge, summarized in Table~\ref{tab:notations}. Given a source domain $\mathcal{S}$ and a target domain $\mathcal{T}$, we use $\st$ to denote the \textit{Cross-domain Performance}, which is the score (e.g., F1) achieved when training a model on data $S\in\mathcal{S}$ and testing it on $T\in\mathcal{T}$. When training and testing the model with data from the source domain, we use $\sss$ to denote the \textit{Source In-domain Performance}. Likewise, $\ttt$ is the \textit{Target In-domain Performance}. 

Finally, we define the \textit{in-domain difference} to be $\idd = \sss - \ttt$. 
A positive $\idd$ may indicate a shift towards an inherently more challenging target domain, for example, the shifts $A \rightarrow C$ and $A \rightarrow B$ from Figure~\ref{fig:intro}. 
The cornerstone of this paper is that \textit{a truthful DR characterization requires considering $\sss$, $\ttt$, and $\st$}. Specifically, full characterization requires understanding the joint distribution of $\sss$, $\ttt$, and $\st$ (see Appendix \S\ref{sec:theorem}). 

Nevertheless, identifying these random variables and their relationships is not tractable without further assumptions, and therefore, we introduce practical and interpretable metrics that quantify the degradation in performance when shifting domains.
We denote the \textit{Average In-domain Performance} by $\overline{\sss} = \E[\sss]$, and the \textit{Average Cross-domain Performance} by $\overline{\st} = \E[\st]$. The difference between these metrics is the \textit{Average Drop}, denoted by $\drop = \overline{\sss} - \overline{\st}$. Intuitively, \textit{the larger the $\drop$ is, the more severe the DR challenge of the model is.}

\subsection{The Source and Target Drops}
\label{sub:drops}

Although characterizing the DR challenge ideally requires task-level analysis across various domain shifts, this approach can be impractical or less relevant when focusing on a specific shift. Hence, we introduce shift-level degradation metrics. The \textit{Source Drop ($\sd$)} and the \textit{Target Drop ($\td$)} are the drops in performance caused by a domain shift, alternately using the source and target's in-domain performance as a point of reference:
\vspace{-0.15cm}
\begin{align*} 
& \sd = \sss - \st \\
& \td = \ttt - \st
\end{align*}
Notice that the training data from the target domain may not be available in a real-life scenario, and in this case, the $\ttt$ can not be computed. The performance degradation we observe in practice is the $\sd$. The $\td$ is a more theoretical measure: 
\textit{``
What would the drop be compared to if the model were trained on data from the target domain?''}

From the above definitions, it follows that: $\sd = \td + \idd$.
This is a solid justification for using both $\sd$ and $\td$ when quantifying the DR challenge. \textit{Using only one could potentially paint an image obscured by the $\idd$, which is not a by-product of the domain shift itself.} For instance, in studies involving challenge sets that report a large $\sd$, the drop may be primarily influenced by a large $\idd$ rather than both $\sd$ and $\td$ being large (e.g., the shift $A \rightarrow C$ in Figure~\ref{fig:intro}). In \S\ref{sub:comparing}, we found that this is the case in many domain shifts.
We refer the readers to \textbf{Appendix \S\ref{sec:theorem} for an extended discussion and theorem} on the relationships between the DR metrics. 

Notably, when a range of experiments is conducted \textbf{using all domains for both training and testing}, it follows that $\E[\sss] = \E[\ttt]$, and from the linearity of the expectation, $\drop = \E[\sd] = \E[\td]$. Importantly, while $\sd$ and $\td$ have equal expected values, they are distinct random variables with differing variances. 
Even though the $\sd$ and the $\td$ are unbiased estimators of the $\drop$, when focusing on specific shifts, such as challenge sets ($\idd>0$), relying solely on a single metric leads to biased conclusions about the robustness of the model. This is exemplified in the toy example of Table~\ref{tab:toy}, and demonstrated in our results in \S\ref{sec:results}.

Finally, other task-level metrics we use are the \textit{Worst $\sd$ ($W_{\sd}$) and Worst $\td$ ($W_{\td}$)}, which measure the highest $\sd$ and $\td$ observed across all domain shifts and identify challenging shifts.

\subsection{Domain Shift Scenarios}
\label{sub:scenarios}

We next introduce a novel framework for classifying domain shifts into four possible scenarios. These scenarios are defined by the sign (positive or negative) of the source and target drops, which can help us understand the nature of the DR challenge. In Appendix \S\ref{sub:intuition}, we further discuss these scenarios and motivate when each might occur.

\medskip\noindent\textbf{The Classic Scenario} ($A \rightarrow B$ in Figure~\ref{fig:intro}) In this scenario both $\sd$ and $\td$ are positive. Accordingly, we deduce that the model is not effectively generalizing from the source domain to the target.

\medskip\noindent\textbf{The Observed Scenario} ($A \rightarrow C$ in Figure~\ref{fig:intro}) This scenario occurs when the shift is to a harder domain and $\ttt < \st < \sss$. In this scenario, only the observed drop, $\sd$ is positive. Although we observe a performance drop, it might be explained by moving to a harder domain and not due to a genuine DR challenge since the model achieves generalization to the target domain and even exhibits higher performance than $\ttt$. 

\medskip\noindent\textbf{The Unobserved Scenario} ($C \rightarrow A$ in Figure~\ref{fig:intro}) This scenario occurs when the shift is from a harder domain to an easier one: $\sss < \st < \ttt$. In this scenario, only $\sd$ is negative, and we do not observe a performance drop. However, since  $\td$ is positive, we know the model can potentially generalize better and it might suffer from a DR challenge. 

\medskip\noindent\textbf{The No Challenge Scenario} ($C \rightarrow B$ in Figure~\ref{fig:intro}) Occurs when $\st$ is larger than both $\sss$ and $\ttt$, therefore, $\sd$ and $\td$ are negative.

\section{The Domain Robustness Benchmark}
\label{sec:benchmark}


\begin{table}
\centering
\large
\begin{adjustbox}{width=0.48\textwidth}
\begin{tabular}{llcccc}
\toprule
\multicolumn{2}{l}{\textbf{Task}} & \textbf{\#D} & \textbf{Train} & \textbf{Dev} &\textbf{Test} \\
\midrule
SA & Sentiment Analysis & 6 & 10K & 2.5K & 2.5K \\ 
NLI & Natural Language Inference & 5 & 50K & 2.5K & 2K \\
AB & Aspect Based SA (ABSA) & 5 & 2K & 500 & 1.4K \\ 
QA & Question Answering & 6 & 9K & 1K & 2.5K \\
QG & Question Generation & 6 & 7.5K & 900 & 1K  \\
AS & Abstractive Summarization & 5 & 10K & 1K & 500 \\
TG & Title Generation & 6 & 17.5K & 1K & 1K  \\
\bottomrule
\end{tabular}
\end{adjustbox}
\caption{Details about the tasks in The Domain Robustness Benchmark. ``\#D'' is the number of domains. ``Train'', ``Dev'', ``Test'' columns present the size of the splits of each domain. Note that we present the average size for the test split since it differs between domains. More details can be found in the project repository.}
\label{tab:dataets}
\vspace{-0.5em}
\end{table}

In Sections \ref{sec:intro} and \ref{sec:related}, we identified shortcomings in existing DR benchmarks. These include an overemphasis on challenge sets and synthetic datasets, coupled with neglecting key NLP tasks such as token-level classification, QA, and particularly generation tasks. To our knowledge, this is the first DR study focusing on various generation tasks, which have gained prominence with the widespread use of LLMs and GenAI. Moreover, most benchmarks consider only a single or very few domains and often use target domains with only test splits, preventing measuring target drops. These limitations restrict a complete understanding of the state of the DR challenge in ``natural settings''. 

To bridge these gaps, we curated a novel DR benchmark that focuses on natural shifts and covers seven downstream tasks. Each task consists of several domains with the same amount of labeled data, enabling using any domain as a source or a target and computing the metrics from \S\ref{sec:methods}. Table~\ref{tab:dataets} details the number of examples in each task domain. In Appendix \S\ref{sec:benchmark_technical}, we describe the preprocessing we performed and discuss technical assumptions. 


\medskip\noindent\textbf{Sentiment Analysis (SA)}
Following \citet{airline_ziser} and \citet{docogen}, we combine five domains of the Amazon product review dataset \citep{sentiment_blitzer} with the airline review dataset \citep{airline} into a single dataset with six domains: \textit{Appliances, Beauty, Books, Games, Software, and Airline}.

\medskip\noindent\textbf{Natural Language Inference (NLI)} 
We use five domains from MNLI dataset \citep{mnli}: \textit{Fiction, Government, Slate, Telephone, and Travel}. 

\medskip\noindent\textbf{Aspect Based Sentiment Analysis (AB)}
Following \citet{dilbert_absa}, we combine the SemEval 2014, 2015, and 2016 \citep{semeval_2014_absa, semeval_2015_absa, semeval_2016_absa} ABSA datasets, together with the MAMs dataset \citep{mams_absa} into a single dataset with four domains: \textit{Device, Laptops, Restaurants, Service, and MAMs}.

\medskip\noindent\textbf{Question Answering (QA)} 
We rely on the SQuAD v2 dataset \citep{squad_v1, squad_v2}, one of the most common QA datasets. We asked human annotators to categorize the documents according to the Wikipedia's taxonomy,\footnote{We merged the vital articles categories: \url{https://en.wikipedia.org/wiki/Wikipedia:Vital_articles}, into eight categories and used six of them as domains.} and created six domains: \textit{Geography, History, Philosophy, Science, Society, and Technology}.

\medskip\noindent\textbf{Question Generation (QG)} 
We rely on our domain partition of the SQuAD dataset \citep{squad_v1} and only use examples with an answer. Given a Wikipedia document and an answer to the question, the task of the NLP model is to generate the question \citep{kd_nlg}.

\medskip\noindent\textbf{Abstractive Summarization (AS)} 
We rely on the Webis-TLDR-17 dataset \citep{reddit_tldr}, which consists of Reddit posts and their ``TL;DR'' summary. We asked human annotators to categorize subreddits 
into five domains: \textit{Drugs, Fitness, LoL (video game), Politics, and Relationships.} 

\medskip\noindent\textbf{Title Generation (TG)} 
We focus on generating titles for Amazon product reviews \citep{titles}. Our dataset contains six domains: \textit{Beauty, Books, DVD, Kitchen, Sports, and Wireless}.

\section{Experimental Setup}
\label{sec:experimental_setup}

\begin{table}
\centering
\normalsize
\begin{adjustbox}{width=0.48\textwidth}
\begin{tabular}{c|lcc|lcc}
\toprule
\textbf{Arch.} & \textbf{Name} & \textbf{\#P} & \textbf{\#L} & \textbf{Name} & \textbf{\#P} & \textbf{\#L} \\
\midrule
\multirowcell{4}{fine-\\tuned\\EO} & \multirowcell{2}[0pt][l]{DistilBert} & \multirowcell{2}{66m} & \multirowcell{2}{6} & DeBERTa-XS & 70m & 12 \\
& & & &  DeBERTa-S & 142m & 6 \\
& RoBERTa-B & 125m & 12 & DeBERTa-B & 184m &  12 \\
& RoBERTa-L & 355m & 24 & DeBERTa-L &  435m & 24 \\
\midrule
\multirowcell{3}{fine-\\tuned\\ED} & T5-S  & 60m & 12 & \multirowcell{2}[0pt][l]{BART-B} & \multirowcell{2}{139m} & \multirowcell{2}{12} \\
& T5-B &  220m & 24 &  & & \\
& T5-L &  737m & 48 & BART-L &  406m & 24 \\
\midrule
\multirowcell{3}{few-\\shot\\DO} & Orca-7b & 7b & 32 & Orca-13b & 13b & 40 \\
& Mistral 7b & 7b & 32 & NeuralChat & 7b & 32 \\
& Llama2-7 & 7b & 32 &  \multirowcell{2}[0pt][l]{Llama2-13b} & \multirowcell{2}[0pt][l]{13b} & \multirowcell{2}[0pt][l]{40} \\
& Llama2-70b & 70b & 40 & & & \\
& GPT3.5 & ? & ? &  GPT4 & ? & ? \\
\bottomrule
\end{tabular}
\end{adjustbox}
\caption{Details about the participating models in this study. 
`Arch.' states the architecture type: EO for Encoder-only, ED for Encoder-decoder, and DO for Decoder-only. `\#P' is the number of parameters in millions (m) or billions (b), and `\#L' is the number of layers.}
\label{tab:models}
\vspace{-0.5em}
\end{table}

Table~\ref{tab:models} presents details about the participating models. Additional implementation details, including hyperparameters and prompts are in Appendix \S\ref{sec:implementation}.

\medskip\noindent\textbf{Fine-tuning Models} For classification tasks (SA, NLI, AB, QA) we employ encoder-only models. Specifically, we use RoBERTa \citep{roberta} and DeBERTa-v3 \citep{deberta_v3}, as well as the smaller DistilBERT \citep{distilbert}. For conditional generation tasks (QG, AS, TG), we utilize two common encoder-decoder models: T5 \citep{t5} and BART \citep{bart}. We chose these open-source models because they offer a variety of sizes (see Table~\ref{tab:models}).

We conduct hyperparameter tuning for each model and source domain, selecting optimal parameters based on the source domain's validation set, and then evaluate the model across all target domains. See Appendix \S\ref{sec:implementation} for more details.

\medskip\noindent\textbf{Zero-shot and Few-shot LLMs} We examine LLMs with an API, including GPT3.5 (\texttt{turbo}) and GPT4 \citep{gpt4}, as well as the open-sourced LLMs LLama v2 \citep{llama_v2}, Orca v2 \citep{orca_v2} (which is based on LLama v2 and fine-tuned using signals from GPT4), 
Mistral-7b \citep{mistral} and NeuralChat \citep{neural_chat} (which is based on Mistral and fine-tuned using the Orca dataset \citep{orca_dataset}). 

For each test example from a target domain, the LLM receives an input comprising a task instruction and the example. In few-shot setups, the input is augmented with additional demonstrations from the source domain. 
Task instructions and prompt examples are provided in Appendix \S\ref{sub:prompts}.

Due to the high costs of API calls and the quadratic increase in the number of experiments with the number of domains,
we limit our presentation of few-shot results to three domains and 600 examples for each task (see Appendix \S\ref{sub:fs_domains}).

\medskip\noindent\textbf{Metrics} For classification tasks (SA, NLI, AB, QA) we report the F1 score. For generation tasks (QG, AS, TG) we report the BertScore \citep{bertscore} with a pre-trained SBERT model \citep{sentence_transformer}. Please see our note in \S\ref{sec:limitations}.L1.

\section{Results}
\label{sec:results}

\begin{table}[t]
\centering
\begin{adjustbox}{width=0.44\textwidth}
\begin{tabular}{l|l|ccccc}
\toprule
\textbf{Task} & \textbf{Model} & {\footnotesize $\overline{\sss}$ } & {\footnotesize $\overline{\st}$ } & {\footnotesize $\drop$ } & {\footnotesize $W_{\sd}$ } & {\footnotesize $W_{\td}$ } \\
\midrule
  \multirowcell{2}{SA} & RoBERTa-L &           95.76 &              92.79 &       2.97 &     13.92 &     19.82 \\
     & DeBERTa-L &           \textbf{96.21} &              \textbf{94.10} &       \textbf{2.11} &      \textbf{9.60} &     \textbf{10.25} \\
\hline
 \multirowcell{2}{NLI} & RoBERTa-L &           89.29 &              87.81 &       \textbf{1.48} &      \textbf{4.83} &      \textbf{2.89} \\
     & DeBERTa-L &           \textbf{90.43} &              \textbf{88.92} &       1.51 &      5.47 &      3.10 \\
\hline
  \multirowcell{2}{AB} & RoBERTa-L &           \textbf{73.31} &              49.42 &      23.90 &     \textbf{35.28} &     \textbf{32.41} \\
     & DeBERTa-L &           71.98 &              \textbf{50.19} &      \textbf{21.80} &     35.54 &     34.49 \\
\hline
  \multirowcell{2}{QA} & RoBERTa-L &           \textbf{82.01} &              \textbf{81.72} &       \textbf{0.29} &      \textbf{6.01} &      \textbf{2.53} \\
     & DeBERTa-L &           74.54 &              74.10 &       0.44 &      6.29 &      2.72 \\
\hline
  \multirowcell{2}{QG} &      T5-L &           \textbf{77.36} &              \textbf{77.24} &       0.13 &      \textbf{4.26} &      1.16 \\
     &    BART-L &           76.30 &              76.30 &      \textbf{0.00} &      4.43 &      \textbf{0.80} \\
\hline
  \multirowcell{2}{AS} &      T5-L &           \textbf{62.40} &              61.42 &       0.98 &      \textbf{4.62} &      2.55 \\
     &    BART-L &           62.33 &              \textbf{61.62} &       \textbf{0.71} &      4.96 &      \textbf{1.93} \\
\hline
  \multirowcell{2}{TG} &      T5-L &           \textbf{66.48} &              \textbf{65.22} &       1.26 &      6.78 &      5.06 \\
     &    BART-L &           65.87 &              64.72 &       \textbf{1.15} &      \textbf{6.61} &      \textbf{4.58} \\
\bottomrule
\end{tabular}
\end{adjustbox}
\caption{Comparison between different large fine-tuned models. The columns are: $\overline{\sss}$ - Average In-domain, $\overline{\st}$ - Average Cross-domain, $\drop$ - Average Drop, $W_{\sd}$ - Worst Source Drop and $W_{\td}$ - Worst Target Drop.}
\label{tab:main_ft}
\vspace{-0.5em}
\end{table}
\begin{table*}[!htb]
\centering
\Large
\begin{adjustbox}{width=1.0\textwidth}
\begin{tabular}{l|cccc|cccc|cccc|cccc|cccc|cccc|cccc}
\toprule
\multirowcell{2}{\textbf{Model}} &  \multicolumn{4}{c|}{\textbf{SA}} & \multicolumn{4}{c|}{\textbf{NLI}} & \multicolumn{4}{c|}{\textbf{AB}} & \multicolumn{4}{c|}{\textbf{QA}} & \multicolumn{4}{c|}{\textbf{QG}} & \multicolumn{4}{c|}{\textbf{AS}} & \multicolumn{4}{c}{\textbf{TG}}  \\ 
& {\normalsize $\overline{\sss}$ } & {\normalsize $\overline{\st}$ } & {\normalsize $W_{\sd}$ } & {\normalsize $W_{\td}$ } & {\normalsize $\overline{\sss}$ } & {\normalsize $\overline{\st}$ } & {\normalsize $W_{\sd}$ } & {\normalsize $W_{\td}$ } & {\normalsize $\overline{\sss}$ } & {\normalsize $\overline{\st}$ } & {\normalsize $W_{\sd}$ } & {\normalsize $W_{\td}$ } & {\normalsize $\overline{\sss}$ } & {\normalsize $\overline{\st}$ } & {\normalsize $W_{\sd}$ } & {\normalsize $W_{\td}$ } & {\normalsize $\overline{\sss}$ } & {\normalsize $\overline{\st}$ } & {\normalsize $W_{\sd}$ } & {\normalsize $W_{\td}$ } & {\normalsize $\overline{\sss}$ } & {\normalsize $\overline{\st}$ } & {\normalsize $W_{\sd}$ } & {\normalsize $W_{\td}$ } & {\normalsize $\overline{\sss}$ } & {\normalsize $\overline{\st}$ } & {\normalsize $W_{\sd}$ } & {\normalsize $W_{\td}$} \\
\midrule
 Orca-7b &              80.9 &                 79.2 & 8.7 &           6.3 &               70.7 &                  70.4 &         16.8 & \textbf{2.5} &              44.0 &                 41.9 &          24.5 &          8.2 &              25.8 &                 23.6 &          5.0 &          3.8 &              65.9 &                 65.5 &          3.0 &          1.5 &              53.4 &                 53.2 &          2.3 &          1.2 &              52.6 &                 52.7 & \textbf{0.6} &          1.0 \\
Orca-13b &              92.6 &                 92.3 &         10.0 &           2.7 &               75.7 &                  74.4 &         15.5 &          6.3 &              52.6 &                 49.2 &          38.3 &         11.9 &              62.8 &                 62.4 &          4.5 &          2.5 &              73.3 &                 73.3 & \textbf{2.2} &          0.8 &              61.0 &                 60.4 &          2.0 &          3.2 &              58.9 &                 58.8 &          1.8 &          0.8 \\
 Mistral &              83.8 &                 80.9 &         11.0 &           5.7 &               49.0 &                  45.8 &         17.1 &         10.3 &              49.9 &                 43.5 &          34.5 &         19.4 &              48.7 &                 46.8 &          6.7 &          7.2 &              69.8 &                 69.5 &          5.2 &          1.2 &              59.0 &                 58.5 &          2.4 &          4.7 &              57.4 &                 57.4 &          1.2 &          1.0 \\
  Neural &              92.4 &                 92.4 &         12.0 &           1.3 &               79.8 &                  77.0 &         16.0 &          8.8 &              42.6 &                 39.3 & \textbf{20.3} &         10.1 &              50.8 &                 49.5 &          5.5 &          4.6 &              72.0 &                 72.1 &          3.9 &          0.8 &              61.6 &                 61.4 & \textbf{1.6} &          1.6 &              58.4 &                 58.3 &          1.6 & \textbf{0.4} \\
 Llama-70b & 94.1 &                  93.9 &          \textbf{8.3} &          1.3 &                56.6 &                   56.3 &           \textbf{5.3} &           4.5 &               51.4 &                  48.6 &         35.9 &          8.9 &               36.6 &                  36.0 &          4.4 &          3.5 &               73.3 &                  73.1 &          4.6 &          0.6 &               60.5 &                  59.2 &          2.7 &          3.5 &               57.7 &                  57.7 &          2.3 &          0.7 \\
  GPT3.5 &              92.1 &                 92.9 &         10.0 & \textbf{0.0} &               72.9 &                  71.7 &         16.4 &          7.2 &              52.7 &        \textbf{51.9} &          37.0 & \textbf{2.4} &              60.1 &                 59.7 &          6.4 &          2.3 &              74.6 &                 74.5 &          4.4 & \textbf{0.3} &     \textbf{64.7} &        \textbf{64.4} &          3.0 &          0.9 &              58.5 &                 58.4 &          1.3 &          0.8 \\
    GPT4 &              95.2 &        \textbf{94.7} &         11.0 &           2.0 &               87.0 &                  86.0 &          6.4 &          3.9 &              51.0 &                 47.9 &          28.9 &          6.9 &              71.0 &                 71.1 &          6.0 & \textbf{0.8} &              76.0 &                 75.8 &          3.5 &          0.5 &              64.1 &                 64.0 &          2.5 & \textbf{0.6} &              58.0 &                 57.9 &          1.5 & \textbf{0.4} \\
\midrule
Best FT &     \textbf{95.5} &                 91.7 &          9.6 &          10.2 &      \textbf{91.0} &         \textbf{89.0} & 5.5 &          3.1 &     \textbf{74.4} &                 47.2 &          35.3 &         32.4 &     \textbf{83.7} &        \textbf{83.5} & \textbf{3.1} &          2.0 &     \textbf{77.7} &        \textbf{77.5} &          4.3 &          0.4 &              63.2 &                 62.0 &          3.5 &          1.7 &     \textbf{65.1} &        \textbf{63.2} &          6.8 &          5.1 \\
\bottomrule
\end{tabular}
\end{adjustbox}
\caption{Comparison between fine-tuned and (4) few-shot models. The `Best FT' selects the best performing fine-tuned model according to the source development set: DeBERTa-L for SA and NLI, RoBERTa-L for AB and QA, and T5-L for QG, AS, TG. All the results are for the same examples and three domains (see Appendix \S\ref{sub:fs_domains}).}
\label{tab:main_fs_results}
\vspace{-0.5em}
\end{table*}
\begin{figure*}[!h]
    \centering
    \includegraphics[width=\textwidth]{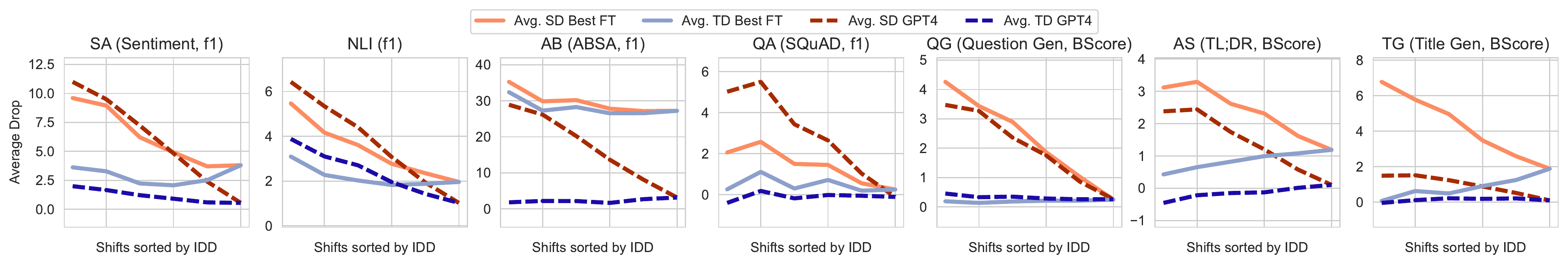}
    \caption{Average $\sd$ (orange lines) and Average $\td$ (blue lines) as a function of challenging domain shifts. Specifically, we sort the domain shifts by their In-domain Difference ($\idd$) and as we move to the right on the x-axis, we incrementally include an additional domain shift in the average drop calculation. Consequently, the leftmost point represents the shift with the largest IDD, while the rightmost point encompasses all shifts. The best fine-tuned model (see caption of Table~\ref{tab:main_fs_results}, solid lines) against GPT4 (dashed lines). This figure illustrates three key findings: (1) The $\sd$ is larger than the $\td$, and when including all shifts their averages are equal; (2) Generally, fine-tuned models exhibit larger drops; (3) Examining only challenging shifts and focusing solely on the $\sd$, obscure the true DR state. Incorporating the $\td$ can compensate for this and provide a clearer understanding.}
    \label{fig:fs_vs_ft}
    \vspace{-0.5em}
\end{figure*}

\subsection{Fine-tuned Models}
\label{sub:fine_tuned_results}

In Table~\ref{tab:main_ft}, we present the results of large fine-tuned models. 
As can be seen, for every task the average in-domain performance consistently exceeds the average cross-domain performance. An exception to this is the QA and QG tasks, which share the same partition of domains, explaining why they behave similarly. Moreover, the vast majority of tasks exhibit non-negligible drops in performance upon domain shift. \textit{This leads to the conclusion that the DR problem still exists in fine-tuned models}, though in varying severity, depending on the task. Some tasks (e.g., AB) exhibit significant drops in most domain shifts, while other tasks (e.g., QA) exhibit minor drops, but we can still expect to have challenging shifts for every domain.

In Appendix \S\ref{sec:additional_results} we provide additional results for fine-tuned models. Specifically, in \S\ref{sub:model_size} we explore the effect of the model size. We observe that larger models 
improve absolute in-domain and cross-domain performance and exhibit an apparent reduction in performance drops, especially in classification tasks. 
In \S\ref{sub:dataset_size}, we examine the impact of the source dataset size. 
We find enhancements in both in-domain and cross-domain performance, however, the performance drop is only reduced in classification tasks and worsens in generation tasks. 


\subsection{Few-shot Models}
\label{sub:few_shot_results}

Unlike fine-tuning, a domain shift occurs for few-shot models when the domain of the prompt demonstrations differs from the test example's domain.  Table~\ref{tab:main_fs_results} presents the results of 4-shots LLMs.  

Similar to fine-tuned models, in most tasks and few-shot models, in-domain performance surpasses cross-domain performance, \textit{indicating that the domain of the demonstration has an effect.} However, the average drops in few-shot models, particularly in GPT3.5 and GPT4, are lower than in fine-tuned models (see also Figure~\ref{fig:fs_vs_ft}). This probably stems from weaker anchoring to the source domain since in few-shot setups, the parameters are not updated based on source domain optimization. Yet, few-shot models experience large worst drops, although, except for NLI and QA, they are much lower than the worst drops of fine-tuned models.

Nevertheless, the robustness of few-shot models comes at a cost of absolute performance. As shown in Table~\ref{tab:main_fs_results}, fine-tuned models outperform all non-GPT models in both in-domain and cross-domain settings. For GPT models, aside from the AS task, the fine-tuned models achieve higher in-domain performance. However, in certain tasks (SA, AB, AS), GPT models exceed the cross-domain performance of fine-tuned models. \textit{This discrepancy highlights the importance of Domain Adaptation research of fine-tuned NLP models.} 

In Appendix \S\ref{sub:number_demonstrations} we study the effect of the number of demonstrations, finding that a larger number of demonstrations usually improves in-domain and cross-domain performance, though in some cases mildly increasing the drop between them (by causing a stronger ``source domain anchoring'').

In Appendix \S\ref{sub:fs_model_size}, we also analyze the impact of few-shot model size. Same as for fine-tuned models, increasing model size generally improves absolute performance and tends to reduce drops.

\begin{figure}[t]
    \centering
    \includegraphics[width=0.48\textwidth]{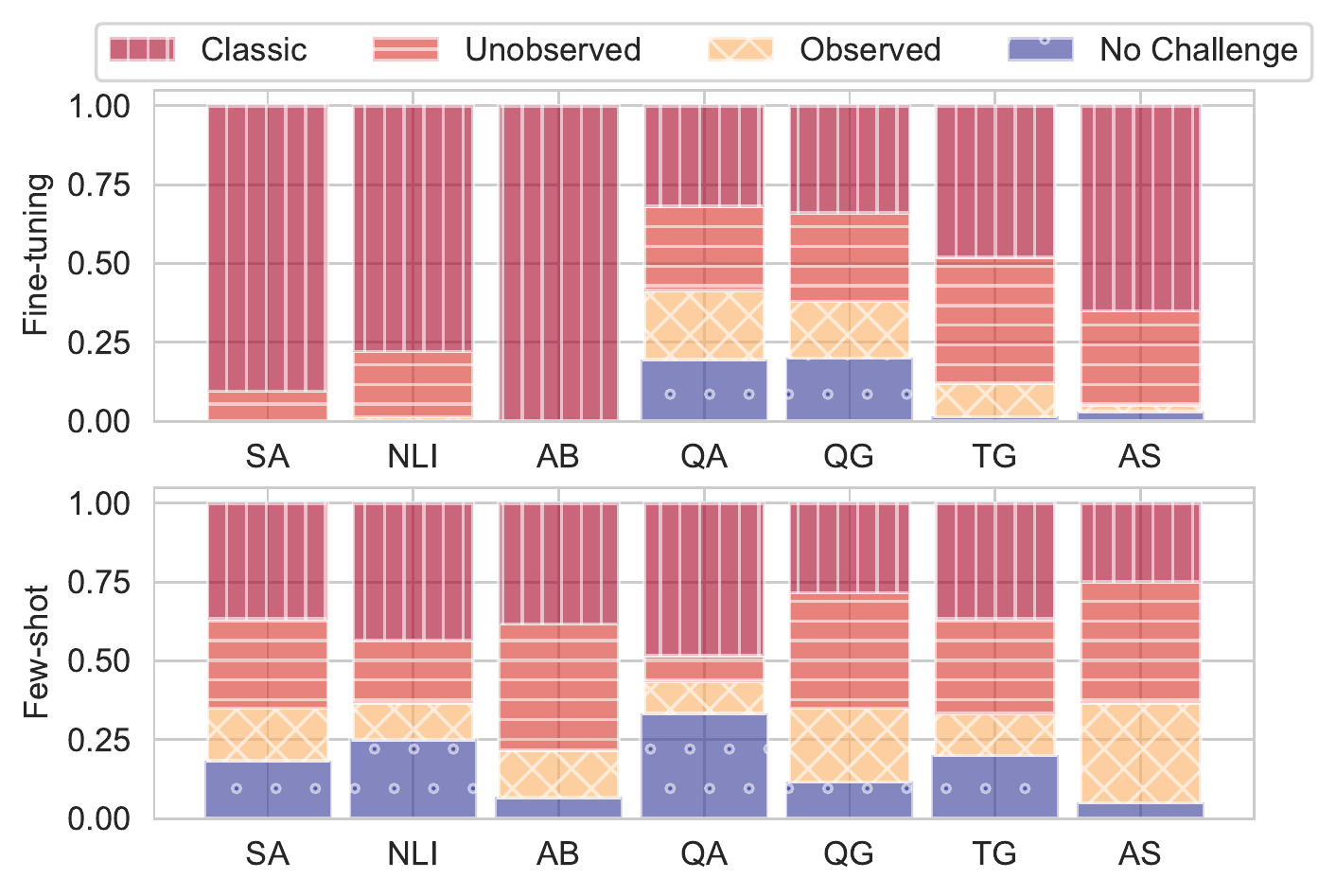}
    \caption{The proportion of each domain shift scenario (see \S\ref{sub:scenarios}) for fine-tuned (top chart) and few-shot models (bottom). For each task, the proportion is measured over all the models and domain shifts. More details in \S\ref{sub:scenarios_stats}.}
    \label{fig:scenarios}
\vspace{-0.8em}
\end{figure}
\begin{table}[!t]
\centering
\footnotesize
\begin{adjustbox}{width=0.48\textwidth}
\begin{tabular}{ll|cc|cc|cc|cc}
\toprule
& \textbf{Task} & \textbf{$\sigma_{\sd}$} & \textbf{$\sigma_{\td}$}
& \textbf{$W_{\sd}$} & \textbf{$W_{\td}$} & \textbf{$\rho_{\sss}$} & \textbf{$\rho_{\ttt}$} & \textbf{$R^2_{\sd}$} & \textbf{$R^2_{\td}$} \\
\midrule
\multirowcell{7}{\rotatebox[origin=c]{90}{Fine-tuning}}      &   SA &    3.62 &    3.33 &     13.23 &     17.05 &     0.28 &     0.42 &     0.34 &     0.08 \\
      &  NLI &    3.06 &    1.29 &      7.14 &      5.12 &    -0.28 &     0.82 &     0.83 &     0.06 \\
      &   AB &    7.09 &    6.53 &     36.05 &     36.55 &    -0.15 &     0.10 &     0.27 &     0.12 \\
      &   QA &    3.71 &    2.07 &      6.76 &      4.52 &    -0.06 &     0.68 &     0.75 &     0.14 \\
      &   QG &    2.29 &    0.46 &      4.55 &      1.21 &    -0.28 &     0.95 &     0.96 &     0.02 \\
      &   AS &    1.91 &    0.65 &      4.81 &      2.49 &     0.06 &     0.77 &     0.95 &     0.58 \\
      &   TG &    2.70 &    1.47 &      6.94 &      4.88 &     0.31 &     0.70 &     0.92 &     0.60 \\
\midrule
\multirowcell{7}{\rotatebox[origin=c]{90}{Few-shot}}      &   SA &    7.49 &    1.77 &     10.58 &      3.73 &    -0.26 &     0.82 &     0.94 &     0.36 \\
      &  NLI &    8.63 &    3.93 &     15.54 &      8.27 &    -0.47 &     0.85 &     0.80 &     0.39 \\
      &   AB &   22.68 &    4.64 &     32.70 &     10.57 &    -0.13 &     0.86 &     0.99 &     0.55 \\
      &   QA &    4.20 &    1.92 &      5.78 &      3.09 &    -0.25 &     0.53 &     0.71 &     0.28 \\
      &   QG &    3.17 &    0.50 &      4.11 &      0.75 &    -0.36 &     0.88 &     0.95 &     0.25 \\
      &   AS &    1.74 &    1.13 &      2.55 &      2.01 &     0.01 &     0.31 &     0.75 &     0.55 \\
      &   TG &    1.21 &    0.55 &      1.62 &      0.91 &    -0.24 &     0.79 &     0.82 &     0.34 \\

\bottomrule
\end{tabular}
\end{adjustbox}
\caption{Statistics of the $\sd$ and the $\td$. We first calculate the statistic for each model and then present the mean statistic for the task. This includes: (1) The standard deviation of the $\sd$ ($\sigma_{\sd}$) and the $\td$ ($\sigma_{\td}$); (2) The Worst $\sd$ ($W_{\sd}$) and $\td$ ($W_{\td}$); (3) Spearman's correlation between the $\st$ and $\sss$ ($\rho_{\sss}$) or $\ttt$ ($\rho_{\ttt}$); (4) The R-squared of $\idd$ and $\sd$ ($R^2_{\sd}$) or $\td$ ($R^2_{\td}$).}
\label{tab:characterizing}
\vspace{-0.5em}
\end{table}

\subsection{Characterizing the DR Challenge}
\label{sub:characterizing}

To understand the nature of the domain shifts, we present the proportion of the four scenarios (from \S\ref{sub:scenarios}) in Figure~\ref{fig:scenarios}. In Appendix~\ref{sub:scenarios_stats}, we provide details on this analysis and confirm its statistical significance. Notably, for fine-tuned models, the Classic scenario, marked by positive $\sd$ and $\td$, emerges as the most dominant and occurs in most tasks with a frequency exceeding 50\%, which indicates \textit{the prevalent DR challenge.} On the other hand, all four scenarios are common across few-shot tasks, suggesting that \textit{the effect of domain shift on few-shot models is weaker and more nuanced.} This is also true in fine-tuned QA and QG tasks, which share the same domain partitions.

Although there is a positive TD in most cases, many are Unobserved scenarios. This finding is essential since many past works overlooked the TD. Our study implies that a DR challenge can exist even when the shift is to an easier domain ($\sss < \ttt$) and even if practitioners do not observe a performance degradation. In comparison, the Observed scenario (positive SD but negative TD), is less frequent but still appears in half of the fine-tuning and few-shot tasks. \textit{This also underscores the necessity for both metrics} and calls for a deeper analysis: which metric more accurately estimates the average drop and cross-domain performance?

\subsection{Comparing SD and TD}
\label{sub:comparing}

In Table~\ref{tab:characterizing}, we see that for every task and for both fine-tuning and few-shot, the variance of the $\sd$ is larger than the variance of the $\td$. In addition, for almost all tasks (except for fine-tuning SA and AB) the Worst $\sd$ is higher than the worst $\td$. These findings indicate that \textit{the $\td$ is a more robust estimator of the average drop.}

Moreover, we find that \textit{the $\st$ behaves more like the $\ttt$ rather than the $\sss$}, as can be seen by Table~\ref{tab:characterizing}, where the correlation between $\st$ and $\ttt$ is much stronger than the correlation with $\sss$ (typically above 0.7). This suggests that attempting to estimate the cross-domain performance without incorporating knowledge of the $\ttt$ is challenging.

Additionally, Table~\ref{tab:characterizing} shows the $R^2$ between the in-domain difference ($\idd = \sss - \ttt$) and the drops. These values indicate the extent to which drop variations can be predicted by the $\idd$. The high $R^2$ of the $\sd$, compared to the $\td$, suggests that \textit{observing a large $\sd$ is likely attributed to shifting to a harder domain and not by genuine DR issues}. This raises a red flag for the NLP community since many works measure DR by source performance degradation on challenge sets. 

The issue becomes clear in Figure~\ref{fig:fs_vs_ft}, which shows the average $\sd$ and $\td$ calculated over challenging shifts. The figure reveals that when focusing on challenging shifts (as shown on the left x-axis), the $\sd$ appears extremely large. Consequently, focusing on challenge sets and relying on the $\sd$ tend to portray a severe picture of the DR state. Examining the $\td$ and additional domain shifts provides a more accurate depiction.

In Appendix \S\ref{sec:theorem}, we provide a detailed discussion of the analysis from this subsection and present a theorem that unifies our findings, demonstrating their equivalence. 
Our study underscores using both metrics, however, when only one is available, the $\td$ is the preferable choice. 

\subsection{Domain Divergence and Drops}
\label{sub:divergence}

\begin{table}[!t]
\centering
\normalsize
\begin{adjustbox}{width=0.48\textwidth}
\begin{tabular}{lc|ccccccc}
\toprule
& & \textbf{SA} & \textbf{NLI}
& \textbf{AB} & \textbf{QA} & \textbf{QG} & \textbf{AS} & \textbf{TG} \\
\midrule
 & $JS-Div$ & 0.23 & 0.32 & 0.27 & 0.30 & 0.27 & 0.18 & 0.18 \\
\midrule
\multirowcell{5}{\rotatebox[origin=c]{90}{Fine-tuning}}      &            $\drop$ &  3.45 & 2.72 & 22.99 &  0.60 &  0.17 &  0.93 &  1.24 \\
   &  $\rho(Div, \sd)$ &  0.43 & 0.02 &  0.53 & -0.02 &  0.02 &  0.07 &  0.09 \\
   &  $\rho(Div, \td)$ &  0.73 & 0.16 &  0.54 &  0.02 &  0.19 &  0.15 &  0.38 \\
   & $\rho(\idd, \sd)$ &  0.53 & 0.91 &  0.51 &  0.86 &  0.98 &  0.98 &  0.96 \\
   & $\rho(\idd, \td)$ & -0.27 & 0.01 & -0.30 & -0.29 & -0.08 & -0.61 & -0.78 \\
\midrule
\multirowcell{5}{\rotatebox[origin=c]{90}{Few-shot}}      &            $\drop$ &  1.29 & 2.20 &  3.37 &  0.49 &  0.13 &  0.45 &  0.18 \\
   & $\rho(Div, \sd)$ &  0.00 & 0.02 &  0.01 & -0.04 &  0.01 &  0.20 &  0.12 \\
   &  $\rho(Div, \td)$ &  0.12 & 0.16 &  0.19 & -0.08 &  0.07 &  0.26 &  0.00 \\
   & $\rho(\idd, \sd)$ &  0.97 & 0.88 &  0.99 &  0.83 &  0.97 &  0.86 &  0.79 \\
   & $\rho(\idd, \td)$ & -0.29 & 0.33 & -0.64 &  0.01 & -0.26 & -0.71 &  0.07 \\
\bottomrule
\end{tabular}
\end{adjustbox}
\caption{Correlations between domain divergence (Jensen-Shannon) and performance drop metrics. 
We first calculate the statistic for each model and then present the mean statistic for the task. The first row presents the average JS-divergence in the task. $\rho(\cdot, \cdot)$ presents the Spearman's correlation. We also present the correlation between the $\idd$ and the performance drop for comparison.}
\label{tab:divergence}
\vspace{-0.5em}
\end{table}

Many past works have explored the connection between domain divergence, a notion of distance between two domains, and the performance drops \citep{remus_divergence, ruder_divergence}. This includes theoretical works that upper-bound the cross-domain performance based on domain divergence \citep{ben_david_divergence, divergence_theory}, and empirical studies that have identified a degree of correlation between divergence metrics and $\sd$ \citep{predict_dr1, divergence_survey}.

While divergence is indeed connected to cross-domain performance and thus to the performance drop, in practice, numerous other factors may influence robustness and performance drops, for example, the $\idd = \sss - \ttt$, which serves to quantify the transition to a more challenging domain and is not a byproduct of a domain shift or a divergence (because it is defined only by $\sss$ and $\ttt$, and not by $\st$). In this subsection, we aim to explore the correlation between domain divergence and the performance drop metrics introduced in this paper.

Following \citet{remus_divergence} and \citet{ruder_divergence}, we decided to use the Jensen Shannon Divergence ($JS\texttt{-}Div$). This decision is based on findings from \citet{divergence_survey}, which demonstrated that, among various divergence metrics, the $JS\texttt{-}Div$ typically shows the highest average correlation. We utilize word frequency distribution to compute the $JS\texttt{-}Div$, excluding stop-words and considering only the top 10k frequent words \citep{divergence_survey}. We then compute for each model and task the correlation between the divergence and the $\sd$ or $\td$ across all pairs of domains. Table~\ref{tab:divergence} presents the average Spearman's correlations.

Our results indicate that stronger correlations between domain divergence and performance drops occur when the DR challenge is more severe. For instance, these correlations are higher for fine-tuned models compared to few-shot models, corresponding with larger average drops ($\drop$). Additionally, we see stronger correlations in tasks such as SA, AB, and TG, which also have larger drops.

In addition, we also present in Table~\ref{tab:divergence} correlations between the $\idd$ and drops. We see that the $\idd$ is a strong predictor (larger magnitude) of the $\sd$, while the opposite holds for domain divergence, which is a better predictor of the $\td$. This is interesting because the domain divergence is theoretically linked to the cross-domain performance, while the $\idd$ is not, further suggesting that the $\td$ is a more reliable estimator of the DR. 

Finally, DR studies typically measure robustness by analyzing shifts only to synthetic, adversarial, or challenge sets, which are known to exhibit high $\idd$. These studies also tend to rely solely on the $\sd$, with high drops suggesting a lack of model robustness. However, our findings raise concerns about the validity of these assessments, which tend to overestimate the severity of the DR challenge, which is generally milder. A more balanced approach would analyze the $\td$ as well, which could help mitigate this bias.

\section{Discussion}
\label{sec:discussion}

In this work, we study the DR challenge in modern NLP models. To this end, we constructed a new DR benchmark comprising various NLP tasks and domain shifts. We proposed shift-level and task-level metrics for precise evaluation and benchmarked numerous fine-tuned models and few-shot LLMs while examining the effect of multiple factors such as model size, dataset size, number of demonstrations, and more. Below, we summarise and discuss the key implications of our work. 

\medskip\noindent\textbf{On Domain Robustness research} 
As discussed in the paper, a full characterization of the DR challenge requires deriving the joint distribution of $\sss$, $\ttt$, and $\st$, which is not tractable. Therefore, we propose practical metrics to quantify the performance degradation: $\sd$ and $\td$. Notably, both metrics are needed for a single domain shift because large drops might be attributed to the in-domain difference ($\idd$) and obscure the DR challenge of the shift.  Indeed, our findings indicate that a large $\sd$ commonly coexists with a large $\idd$. 

At the task level, the expected values of both drop metrics ($\sd$ and $\td$) are equal and correspond to the average drop ($\drop$, see \S\ref{sec:theorem}). We found empirically that the $\td$ is a better estimator of the $\drop$. This implies that when examining a limited number of shifts, including the $\td$ is crucial. 

Moreover, we suggest that current research may paint an inaccurate picture of the state of domain robustness. This stems from two of our findings. First, performance degradation is larger when measured with the $\sd$ than with the $\td$. Second, every task has shifts with severe performance drops, even when most shifts are not remotely as bad. 
Past works, which typically focused on challenging shifts such as challenge sets ($\idd > 0$) and rely solely on the $\sd$, present a distorted image of the actual state of DR, which is actually much milder.

\medskip\noindent\textbf{On predicting cross-domain performance}
Estimating model performance has an important impact on the deployment and maintenance of NLP models and related financial decisions, such as the need for annotation \citep{predict_dr2, predict_dr1, predict_no_asnwer, al_da}. 
Our findings suggest that the $\ttt$ is a better predictor of the cross-domain performance ($\st$) than the $\sss$. Accordingly, knowledge about the target domain (such as a sample of labeled data) is essential, and without it, estimators may struggle to predict cross-domain performance.
In addition, previous studies have attempted to predict performance drops (only the $\sd$) using domain divergence \citep{divergence_survey}. Our study suggests that domain divergence is a better predictor of the $\td$ than the $\sd$.

\medskip\noindent\textbf{On the relevance of fine-tuning} 
While zero-shot and few-shot LLMs can perform various tasks without the additional cost of annotating data or training a model, we still believe that fine-tuning a smaller model remains the de-facto standard \citep{standing_on}. First, LLM usage can be very costly, as LLMs require massive computational resources and extremely high latency. Second, LLMs that cannot be fine-tuned may be less effective when the data cannot be sent to external servers because of privacy constraints or when the domain is unique or specific (e.g., in national security settings or human conversations).
Third, with enough task-specific labeled data that few-shot LLMs can cheaply annotate, it is possible to develop a small, high-performing, fine-tuned model \citep{kd_nlg, faithful_exp, TrueTeacher, comblm}.  
Fourth, there is strong evidence that few-shot language models underperform fine-tuned models in specific domains that require expertise, such as biomedical \citep{gpt3_biomedical} or when the training size is large  \citep{revisiting}. This study shows that task-specific fine-tuned models outperform few-shot models in-domain, although this gap may be closed soon. 

Nevertheless, we also found that few-shot LLMs are more robust to domain shifts and can outperform fine-tuned models cross-domain. This calls for further domain adaptation research on fine-tuned models. 

\medskip\noindent\textbf{On the relevance of Domain Adaptation} 
Domain Adaptation (DA) is a field that addresses solutions to the DR problem. DA research considers various setups, each having different assumptions on the availability of data from the target domain at the model training time \citep{sentiment_blitzer, da_data_selection, ziser1, ziser3, self_training_rotman, perl, unsupervised_da_survey, adapters, pada, docogen, hyperpada, da_prompts, da_la23, da_li23, da_va23}. 

Modern NLP models are believed to be robust due to the pretraining process, where the models have seen a vast amount of diverse data from various domains. Another reason could be data contamination \citep{contamination, contamination2}, i.e., pretraining on data from a downstream task improves the performance on it \citep{gpt2, dg_continue_pretraining_classification, continue_pretraining_for_da}. This belief questions the relevance or the necessity of Domain Adaptation research.

However, in this study, we demonstrated that the DR challenge still exists. We show that there is a performance drop due to domain shift in every task or model, and some shifts are remarkably challenging. We believe that DA research remains essential and relevant, particularly for NLP. 
To facilitate further research, we provide an NLP benchmark with natural topic shifts, which has some challenging setups for various NLP tasks. We hope this benchmark will be used to evaluate and improve DA methods.

\section{Limitations}
\label{sec:limitations}

\medskip\noindent\textbf{L1. Prompt Engineering} 
Noteworthy, we experimented with various prompts and task instruction revisions but saw no significant change. Following \citet{few_shot_demonstrations}, we also tried selecting demonstrations from the source domain most similar to the target test example using a pre-trained SBERT model \citep{sentence_transformer}. However, this approach did not enhance performance and introduced biases, such as demonstrations from only one class, leading the LLM to classify the test example with this class. 


\medskip\noindent\textbf{L2. Additional Domain Shifts} 
In developing our benchmark, we grounded it in technical assumptions aimed at facilitating a controlled experimental analysis, as detailed \S\ref{sub:benchmark_assumptions}. One of these assumptions is to focus on natural topic shifts (although other factors are likely to change as well, such as the style and syntax). This contrasts with other studies that explore synthetic shifts, such as adversarial attacks, challenge sets, or transitions to datasets from different data-generating processes (e.g., having other annotation guidelines). 
Our rationale for this approach was to isolate and control a single variable, thus allowing for a more scientific and precise characterization of the DR challenge. 

While our assumptions simplify the domain shift, we argue that if the DR challenge exists under these assumptions (and it does), then it will definitely exist more severely when our assumptions are violated and a complex shift occurs. Researchers who wish to focus on a specific type of prior shift (e.g., unbalanced domains) can easily use our benchmark to construct more challenging setups.

\medskip\noindent\textbf{L3. Domain Adaptation Solutions} 
Although a wide array of DA solutions exists to address the DR challenge and improve the OOD generalization of NLP models, our study specifically focuses on the diagnostic aspect. We aim to explore whether this challenge is prevalent in modern NLP models, and our findings confirm it is prevalent. We anticipate that future research could leverage our new DR benchmark for diagnostic purposes as well as for benchmarking DA solutions. Furthermore, we hope our study will facilitate further research in this vital area and inspire novel DA methods.

\medskip\noindent\textbf{L4. Text Generation Evaluation} 
Text generation evaluation is an open research problem, and many techniques exist. Although we report BERTScore for simplicity, we did conduct a comprehensive analysis using various metrics (BLEU, ROUGE, METEOR, BLEURT, etc...) and observed similar trends to our findings. We chose BERTScore because it captures semantic similarity and context. In addition, upon manual inspection of LLM outputs, we found them comparable or even superior to the reference texts used for benchmarking. Yet, automatic evaluation with references is useful for assessing the extent to which models learn and capture the dataset distribution $P(Y|X)$. This perspective shifts the focus from human preference to a more technical objective. Supporting this viewpoint is the fact that increasing the number of demonstrations also enhances the performance.



\bibliography{custom}

\begin{thebibliography}{94}
\expandafter\ifx\csname natexlab\endcsname\relax\def\natexlab#1{#1}\fi

\bibitem[{Belinkov and Bisk(2018)}]{synthetic_nmt}
Yonatan Belinkov and Yonatan Bisk. 2018.
\newblock \href {https://openreview.net/forum?id=BJ8vJebC-} {Synthetic and natural noise both break neural machine translation}.
\newblock In \emph{6th International Conference on Learning Representations, {ICLR} 2018, Vancouver, BC, Canada, April 30 - May 3, 2018, Conference Track Proceedings}. OpenReview.net.

\bibitem[{Ben{-}David et~al.(2022{\natexlab{a}})Ben{-}David, Oved, and Reichart}]{pada}
Eyal Ben{-}David, Nadav Oved, and Roi Reichart. 2022{\natexlab{a}}.
\newblock \href {https://doi.org/10.1162/tacl\_a\_00468} {{PADA:} example-based prompt learning for on-the-fly adaptation to unseen domains}.
\newblock \emph{Trans. Assoc. Comput. Linguistics}, 10:414--433.

\bibitem[{Ben{-}David et~al.(2020)Ben{-}David, Rabinovitz, and Reichart}]{perl}
Eyal Ben{-}David, Carmel Rabinovitz, and Roi Reichart. 2020.
\newblock \href {https://doi.org/10.1162/tacl\_a\_00328} {{PERL:} pivot-based domain adaptation for pre-trained deep contextualized embedding models}.
\newblock \emph{Trans. Assoc. Comput. Linguistics}, 8:504--521.

\bibitem[{Ben{-}David et~al.(2022{\natexlab{b}})Ben{-}David, Ziser, and Reichart}]{al_da}
Eyal Ben{-}David, Yftah Ziser, and Roi Reichart. 2022{\natexlab{b}}.
\newblock \href {https://doi.org/10.48550/arXiv.2209.00830} {Domain adaptation from scratch}.
\newblock \emph{CoRR}, abs/2209.00830.

\bibitem[{Ben{-}David et~al.(2010)Ben{-}David, Blitzer, Crammer, Kulesza, Pereira, and Vaughan}]{ben_david_divergence}
Shai Ben{-}David, John Blitzer, Koby Crammer, Alex Kulesza, Fernando Pereira, and Jennifer~Wortman Vaughan. 2010.
\newblock \href {https://doi.org/10.1007/S10994-009-5152-4} {A theory of learning from different domains}.
\newblock \emph{Mach. Learn.}, 79(1-2):151--175.

\bibitem[{Blitzer et~al.(2007)Blitzer, Dredze, and Pereira}]{sentiment_blitzer}
John Blitzer, Mark Dredze, and Fernando Pereira. 2007.
\newblock \href {https://aclanthology.org/P07-1056/} {Biographies, bollywood, boom-boxes and blenders: Domain adaptation for sentiment classification}.
\newblock In \emph{{ACL} 2007, Proceedings of the 45th Annual Meeting of the Association for Computational Linguistics, June 23-30, 2007, Prague, Czech Republic}. The Association for Computational Linguistics.

\bibitem[{Brown et~al.(2020)Brown, Mann, Ryder, Subbiah, and Jared~Kaplan}]{gpt3}
Tom~B. Brown, Benjamin Mann, Nick Ryder, Melanie Subbiah, and et~al. Jared~Kaplan. 2020.
\newblock \href {https://proceedings.neurips.cc/paper/2020/hash/1457c0d6bfcb4967418bfb8ac142f64a-Abstract.html} {Language models are few-shot learners}.
\newblock In \emph{Advances in Neural Information Processing Systems 33: Annual Conference on Neural Information Processing Systems 2020, NeurIPS 2020, December 6-12, 2020, virtual}.

\bibitem[{Budzianowski et~al.(2018)Budzianowski, Wen, Tseng, Casanueva, Ultes, Ramadan, and Gasic}]{multi_woz}
Pawel Budzianowski, Tsung{-}Hsien Wen, Bo{-}Hsiang Tseng, I{\~{n}}igo Casanueva, Stefan Ultes, Osman Ramadan, and Milica Gasic. 2018.
\newblock \href {https://aclanthology.org/D18-1547/} {Multiwoz - {A} large-scale multi-domain wizard-of-oz dataset for task-oriented dialogue modelling}.
\newblock In \emph{Proceedings of the 2018 Conference on Empirical Methods in Natural Language Processing, Brussels, Belgium, October 31 - November 4, 2018}, pages 5016--5026. Association for Computational Linguistics.

\bibitem[{Calderon et~al.(2022)Calderon, Ben{-}David, Feder, and Reichart}]{docogen}
Nitay Calderon, Eyal Ben{-}David, Amir Feder, and Roi Reichart. 2022.
\newblock \href {https://doi.org/10.18653/v1/2022.acl-long.533} {Docogen: Domain counterfactual generation for low resource domain adaptation}.
\newblock In \emph{Proceedings of the 60th Annual Meeting of the Association for Computational Linguistics (Volume 1: Long Papers), {ACL} 2022, Dublin, Ireland, May 22-27, 2022}, pages 7727--7746. Association for Computational Linguistics.

\bibitem[{Calderon et~al.(2023)Calderon, Mukherjee, Reichart, and Kantor}]{kd_nlg}
Nitay Calderon, Subhabrata Mukherjee, Roi Reichart, and Amir Kantor. 2023.
\newblock \href {https://doi.org/10.48550/arXiv.2305.02031} {A systematic study of knowledge distillation for natural language generation with pseudo-target training}.
\newblock \emph{CoRR}, abs/2305.02031.

\bibitem[{Chalkidis et~al.(2020)Chalkidis, Fergadiotis, Malakasiotis, Aletras, and Androutsopoulos}]{dg_example_legal}
Ilias Chalkidis, Manos Fergadiotis, Prodromos Malakasiotis, Nikolaos Aletras, and Ion Androutsopoulos. 2020.
\newblock \href {http://arxiv.org/abs/2010.02559} {{LEGAL-BERT:} the muppets straight out of law school}.
\newblock \emph{CoRR}, abs/2010.02559.

\bibitem[{Chowdhery et~al.(2022)Chowdhery, Narang, Devlin, Bosma, Mishra, and ~}]{palm}
Aakanksha Chowdhery, Sharan Narang, Jacob Devlin, Maarten Bosma, Gaurav Mishra, and et.~al. ~. 2022.
\newblock \href {https://doi.org/10.48550/arXiv.2204.02311} {Palm: Scaling language modeling with pathways}.
\newblock \emph{CoRR}, abs/2204.02311:30.

\bibitem[{Chronopoulou et~al.(2022)Chronopoulou, Peters, and Dodge}]{lm_da_benchmark_2}
Alexandra Chronopoulou, Matthew~E. Peters, and Jesse Dodge. 2022.
\newblock \href {https://doi.org/10.18653/v1/2022.naacl-main.96} {Efficient hierarchical domain adaptation for pretrained language models}.
\newblock In \emph{Proceedings of the 2022 Conference of the North American Chapter of the Association for Computational Linguistics: Human Language Technologies, {NAACL} 2022, Seattle, WA, United States, July 10-15, 2022}, pages 1336--1351. Association for Computational Linguistics.

\bibitem[{Cvejoski et~al.(2022)Cvejoski, S{\'{a}}nchez, and Ojeda}]{robustness_reddit_popularity_prediction}
Kostadin Cvejoski, Rams{\'{e}}s~J. S{\'{a}}nchez, and C{\'{e}}sar Ojeda. 2022.
\newblock \href {https://doi.org/10.48550/arXiv.2211.00384} {The future is different: Large pre-trained language models fail in prediction tasks}.
\newblock \emph{CoRR}, abs/2211.00384.

\bibitem[{Devlin et~al.(2018)Devlin, Chang, Lee, and Toutanova}]{bert}
Jacob Devlin, Ming{-}Wei Chang, Kenton Lee, and Kristina Toutanova. 2018.
\newblock \href {http://arxiv.org/abs/1810.04805} {{BERT:} pre-training of deep bidirectional transformers for language understanding}.
\newblock \emph{CoRR}, abs/1810.04805.

\bibitem[{ElSahar and Gall{\'{e}}(2019)}]{predict_dr1}
Hady ElSahar and Matthias Gall{\'{e}}. 2019.
\newblock \href {https://doi.org/10.18653/v1/D19-1222} {To annotate or not? predicting performance drop under domain shift}.
\newblock In \emph{Proceedings of the 2019 Conference on Empirical Methods in Natural Language Processing and the 9th International Joint Conference on Natural Language Processing, {EMNLP-IJCNLP} 2019, Hong Kong, China, November 3-7, 2019}, pages 2163--2173. Association for Computational Linguistics.

\bibitem[{Gao et~al.(2021)Gao, Fisch, and Chen}]{few_shot_demonstrations}
Tianyu Gao, Adam Fisch, and Danqi Chen. 2021.
\newblock \href {https://doi.org/10.18653/v1/2021.acl-long.295} {Making pre-trained language models better few-shot learners}.
\newblock In \emph{Proceedings of the 59th Annual Meeting of the Association for Computational Linguistics and the 11th International Joint Conference on Natural Language Processing, {ACL/IJCNLP} 2021, (Volume 1: Long Papers), Virtual Event, August 1-6, 2021}, pages 3816--3830. Association for Computational Linguistics.

\bibitem[{Gat et~al.(2023)Gat, Calderon, Feder, Chapanin, Sharma, and Reichart}]{faithful_exp}
Yair~Ori Gat, Nitay Calderon, Amir Feder, Alexander Chapanin, Amit Sharma, and Roi Reichart. 2023.
\newblock \href {https://doi.org/10.48550/ARXIV.2310.00603} {Faithful explanations of black-box {NLP} models using llm-generated counterfactuals}.
\newblock \emph{CoRR}, abs/2310.00603.

\bibitem[{Ge et~al.(2023)Ge, Huang, Xie, Lai, Song, Li, and Huang}]{da_prompts}
Chunjiang Ge, Rui Huang, Mixue Xie, Zihang Lai, Shiji Song, Shuang Li, and Gao Huang. 2023.
\newblock \href {https://doi.org/10.1109/TNNLS.2023.3327962} {Domain adaptation via prompt learning}.
\newblock \emph{IEEE Transactions on Neural Networks and Learning Systems}, pages 1--11.

\bibitem[{Gekhman et~al.(2023{\natexlab{a}})Gekhman, Herzig, Aharoni, Elkind, and Szpektor}]{TrueTeacher}
Zorik Gekhman, Jonathan Herzig, Roee Aharoni, Chen Elkind, and Idan Szpektor. 2023{\natexlab{a}}.
\newblock \href {https://aclanthology.org/2023.emnlp-main.127} {Trueteacher: Learning factual consistency evaluation with large language models}.
\newblock In \emph{Proceedings of the 2023 Conference on Empirical Methods in Natural Language Processing, {EMNLP} 2023, Singapore, December 6-10, 2023}, pages 2053--2070. Association for Computational Linguistics.

\bibitem[{Gekhman et~al.(2023{\natexlab{b}})Gekhman, Oved, Keller, Szpektor, and Reichart}]{MarCQAp}
Zorik Gekhman, Nadav Oved, Orgad Keller, Idan Szpektor, and Roi Reichart. 2023{\natexlab{b}}.
\newblock \href {https://doi.org/10.1162/tacl_a_00549} {On the robustness of dialogue history representation in conversational question answering: A comprehensive study and a new prompt-based method}.
\newblock \emph{Transactions of the Association for Computational Linguistics}, 11:351--366.

\bibitem[{Goel et~al.(2021)Goel, Rajani, Vig, Taschdjian, Bansal, and R{\'{e}}}]{dr_gym}
Karan Goel, Nazneen~Fatema Rajani, Jesse Vig, Zachary Taschdjian, Mohit Bansal, and Christopher R{\'{e}}. 2021.
\newblock \href {https://doi.org/10.18653/V1/2021.NAACL-DEMOS.6} {Robustness gym: Unifying the {NLP} evaluation landscape}.
\newblock In \emph{Proceedings of the 2021 Conference of the North American Chapter of the Association for Computational Linguistics: Human Language Technologies: Demonstrations, {NAACL-HLT} 2021, Online, June 6-11, 2021}, pages 42--55. Association for Computational Linguistics.

\bibitem[{Gururangan et~al.(2020)Gururangan, Marasovic, Swayamdipta, Lo, Beltagy, Downey, and Smith}]{continue_pretraining_for_da}
Suchin Gururangan, Ana Marasovic, Swabha Swayamdipta, Kyle Lo, Iz~Beltagy, Doug Downey, and Noah~A. Smith. 2020.
\newblock \href {https://doi.org/10.18653/v1/2020.acl-main.740} {Don't stop pretraining: Adapt language models to domains and tasks}.
\newblock In \emph{Proceedings of the 58th Annual Meeting of the Association for Computational Linguistics, {ACL} 2020, Online, July 5-10, 2020}, pages 8342--8360. Association for Computational Linguistics.

\bibitem[{Gutierrez et~al.(2022)Gutierrez, McNeal, Washington, Chen, Li, Sun, and Su}]{gpt3_biomedical}
Bernal~Jimenez Gutierrez, Nikolas McNeal, Clayton Washington, You Chen, Lang Li, Huan Sun, and Yu~Su. 2022.
\newblock \href {https://aclanthology.org/2022.findings-emnlp.329} {Thinking about {GPT-3} in-context learning for biomedical ie? think again}.
\newblock In \emph{Findings of the Association for Computational Linguistics: {EMNLP} 2022, Abu Dhabi, United Arab Emirates, December 7-11, 2022}, pages 4497--4512. Association for Computational Linguistics.

\bibitem[{Han and Eisenstein(2019)}]{dg_continue_pretraining_classification}
Xiaochuang Han and Jacob Eisenstein. 2019.
\newblock \href {https://doi.org/10.18653/v1/D19-1433} {Unsupervised domain adaptation of contextualized embeddings for sequence labeling}.
\newblock In \emph{Proceedings of the 2019 Conference on Empirical Methods in Natural Language Processing and the 9th International Joint Conference on Natural Language Processing, {EMNLP-IJCNLP} 2019, Hong Kong, China, November 3-7, 2019}, pages 4237--4247. Association for Computational Linguistics.

\bibitem[{He et~al.(2021{\natexlab{a}})He, Gao, and Chen}]{deberta_v3}
Pengcheng He, Jianfeng Gao, and Weizhu Chen. 2021{\natexlab{a}}.
\newblock \href {http://arxiv.org/abs/2111.09543} {Debertav3: Improving deberta using electra-style pre-training with gradient-disentangled embedding sharing}.
\newblock \emph{CoRR}, abs/2111.09543.

\bibitem[{He et~al.(2021{\natexlab{b}})He, Liu, Ye, Tan, Ding, Cheng, Low, Bing, and Si}]{adapters}
Ruidan He, Linlin Liu, Hai Ye, Qingyu Tan, Bosheng Ding, Liying Cheng, Jia{-}Wei Low, Lidong Bing, and Luo Si. 2021{\natexlab{b}}.
\newblock \href {https://doi.org/10.18653/v1/2021.acl-long.172} {On the effectiveness of adapter-based tuning for pretrained language model adaptation}.
\newblock In \emph{Proceedings of the 59th Annual Meeting of the Association for Computational Linguistics and the 11th International Joint Conference on Natural Language Processing, {ACL/IJCNLP} 2021, (Volume 1: Long Papers), Virtual Event, August 1-6, 2021}, pages 2208--2222. Association for Computational Linguistics.

\bibitem[{Hendrycks et~al.(2020)Hendrycks, Liu, Wallace, Dziedzic, Krishnan, and Song}]{transformers_improve_robustness}
Dan Hendrycks, Xiaoyuan Liu, Eric Wallace, Adam Dziedzic, Rishabh Krishnan, and Dawn Song. 2020.
\newblock \href {https://doi.org/10.18653/v1/2020.acl-main.244} {Pretrained transformers improve out-of-distribution robustness}.
\newblock In \emph{Proceedings of the 58th Annual Meeting of the Association for Computational Linguistics, {ACL} 2020, Online, July 5-10, 2020}, pages 2744--2751. Association for Computational Linguistics.

\bibitem[{Hupkes et~al.(2023)Hupkes, Giulianelli, Dankers, Artetxe, Elazar, Pimentel, Christodoulopoulos, Lasri, Saphra, Sinclair, Ulmer, Schottmann, Batsuren, Sun, Sinha, Khalatbari, Ryskina, Frieske, Cotterell, and Jin}]{da_hu23}
Dieuwke Hupkes, Mario Giulianelli, Verna Dankers, Mikel Artetxe, Yanai Elazar, Tiago Pimentel, Christos~E. Christodoulopoulos, Karim Lasri, Naomi Saphra, Arabella Sinclair, Dennis Ulmer, Florian Schottmann, Khuyagbaatar Batsuren, Kaiser Sun, Koustuv Sinha, Leila Khalatbari, Maria Ryskina, Rita Frieske, Ryan Cotterell, and Zhijing Jin. 2023.
\newblock \href {https://doi.org/10.1038/S42256-023-00729-Y} {A taxonomy and review of generalization research in {NLP}}.
\newblock \emph{Nat. Mac. Intell.}, 5(10):1161--1174.

\bibitem[{Jiang et~al.(2023)Jiang, Sablayrolles, Mensch, Bamford, Chaplot, de~Las~Casas, Bressand, Lengyel, Lample, Saulnier, Lavaud, Lachaux, Stock, Scao, Lavril, Wang, Lacroix, and Sayed}]{mistral}
Albert~Q. Jiang, Alexandre Sablayrolles, Arthur Mensch, Chris Bamford, Devendra~Singh Chaplot, Diego de~Las~Casas, Florian Bressand, Gianna Lengyel, Guillaume Lample, Lucile Saulnier, L{\'{e}}lio~Renard Lavaud, Marie{-}Anne Lachaux, Pierre Stock, Teven~Le Scao, Thibaut Lavril, Thomas Wang, Timoth{\'{e}}e Lacroix, and William~El Sayed. 2023.
\newblock \href {https://doi.org/10.48550/ARXIV.2310.06825} {Mistral 7b}.
\newblock \emph{CoRR}, abs/2310.06825.

\bibitem[{Jiang et~al.(2019)Jiang, Chen, Xu, Ao, and Yang}]{mams_absa}
Qingnan Jiang, Lei Chen, Ruifeng Xu, Xiang Ao, and Min Yang. 2019.
\newblock \href {https://doi.org/10.18653/v1/D19-1654} {A challenge dataset and effective models for aspect-based sentiment analysis}.
\newblock In \emph{Proceedings of the 2019 Conference on Empirical Methods in Natural Language Processing and the 9th International Joint Conference on Natural Language Processing, {EMNLP-IJCNLP} 2019, Hong Kong, China, November 3-7, 2019}, pages 6279--6284. Association for Computational Linguistics.

\bibitem[{Jin et~al.(2020)Jin, Jin, Zhou, and Szolovits}]{adversarial_bert}
Di~Jin, Zhijing Jin, Joey~Tianyi Zhou, and Peter Szolovits. 2020.
\newblock \href {https://ojs.aaai.org/index.php/AAAI/article/view/6311} {Is {BERT} really robust? {A} strong baseline for natural language attack on text classification and entailment}.
\newblock In \emph{The Thirty-Fourth {AAAI} Conference on Artificial Intelligence, {AAAI} 2020, The Thirty-Second Innovative Applications of Artificial Intelligence Conference, {IAAI} 2020, The Tenth {AAAI} Symposium on Educational Advances in Artificial Intelligence, {EAAI} 2020, New York, NY, USA, February 7-12, 2020}, pages 8018--8025. {AAAI} Press.

\bibitem[{Kashyap et~al.(2021)Kashyap, Hazarika, Kan, and Zimmermann}]{divergence_survey}
Abhinav~Ramesh Kashyap, Devamanyu Hazarika, Min{-}Yen Kan, and Roger Zimmermann. 2021.
\newblock \href {https://doi.org/10.18653/V1/2021.NAACL-MAIN.147} {Domain divergences: {A} survey and empirical analysis}.
\newblock In \emph{Proceedings of the 2021 Conference of the North American Chapter of the Association for Computational Linguistics: Human Language Technologies, {NAACL-HLT} 2021, Online, June 6-11, 2021}, pages 1830--1849. Association for Computational Linguistics.

\bibitem[{Kaushik et~al.(2020)Kaushik, Hovy, and Lipton}]{counterfactual_set}
Divyansh Kaushik, Eduard~H. Hovy, and Zachary~Chase Lipton. 2020.
\newblock \href {https://openreview.net/forum?id=Sklgs0NFvr} {Learning the difference that makes {A} difference with counterfactually-augmented data}.
\newblock In \emph{8th International Conference on Learning Representations, {ICLR} 2020, Addis Ababa, Ethiopia, April 26-30, 2020}. OpenReview.net.

\bibitem[{Koh et~al.(2021)Koh, Sagawa, Marklund, Xie, Zhang, Balsubramani, Hu, Yasunaga, Phillips, Gao, Lee, David, Stavness, Guo, Earnshaw, Haque, Beery, Leskovec, Kundaje, Pierson, Levine, Finn, and Liang}]{wilds}
Pang~Wei Koh, Shiori Sagawa, Henrik Marklund, Sang~Michael Xie, Marvin Zhang, Akshay Balsubramani, Weihua Hu, Michihiro Yasunaga, Richard~Lanas Phillips, Irena Gao, Tony Lee, Etienne David, Ian Stavness, Wei Guo, Berton Earnshaw, Imran~S. Haque, Sara~M. Beery, Jure Leskovec, Anshul Kundaje, Emma Pierson, Sergey Levine, Chelsea Finn, and Percy Liang. 2021.
\newblock \href {http://proceedings.mlr.press/v139/koh21a.html} {{WILDS:} {A} benchmark of in-the-wild distribution shifts}.
\newblock In \emph{Proceedings of the 38th International Conference on Machine Learning, {ICML} 2021, 18-24 July 2021, Virtual Event}, volume 139 of \emph{Proceedings of Machine Learning Research}, pages 5637--5664. {PMLR}.

\bibitem[{Lang et~al.(2023)Lang, Zheng, Li, Sun, Huang, and Li}]{da_la23}
Hao Lang, Yinhe Zheng, Yixuan Li, Jian Sun, Fei Huang, and Yongbin Li. 2023.
\newblock \href {https://doi.org/10.48550/ARXIV.2305.03236} {A survey on out-of-distribution detection in {NLP}}.
\newblock \emph{CoRR}, abs/2305.03236.

\bibitem[{Lekhtman et~al.(2021)Lekhtman, Ziser, and Reichart}]{dilbert_absa}
Entony Lekhtman, Yftah Ziser, and Roi Reichart. 2021.
\newblock \href {https://doi.org/10.18653/v1/2021.emnlp-main.20} {{DILBERT:} customized pre-training for domain adaptation with category shift, with an application to aspect extraction}.
\newblock In \emph{Proceedings of the 2021 Conference on Empirical Methods in Natural Language Processing, {EMNLP} 2021, Virtual Event / Punta Cana, Dominican Republic, 7-11 November, 2021}, pages 219--230. Association for Computational Linguistics.

\bibitem[{Levine et~al.(2022)Levine, Dalmedigos, Ram, Zeldes, Jannai, Muhlgay, Osin, Lieber, Lenz, Shalev{-}Shwartz, Shashua, Leyton{-}Brown, and Shoham}]{standing_on}
Yoav Levine, Itay Dalmedigos, Ori Ram, Yoel Zeldes, Daniel Jannai, Dor Muhlgay, Yoni Osin, Opher Lieber, Barak Lenz, Shai Shalev{-}Shwartz, Amnon Shashua, Kevin Leyton{-}Brown, and Yoav Shoham. 2022.
\newblock \href {https://doi.org/10.48550/arXiv.2204.10019} {Standing on the shoulders of giant frozen language models}.
\newblock \emph{CoRR}, abs/2204.10019.

\bibitem[{Lewis et~al.(2020)Lewis, Liu, Goyal, Ghazvininejad, Mohamed, Levy, Stoyanov, and Zettlemoyer}]{bart}
Mike Lewis, Yinhan Liu, Naman Goyal, Marjan Ghazvininejad, Abdelrahman Mohamed, Omer Levy, Veselin Stoyanov, and Luke Zettlemoyer. 2020.
\newblock \href {https://doi.org/10.18653/v1/2020.acl-main.703} {{BART:} denoising sequence-to-sequence pre-training for natural language generation, translation, and comprehension}.
\newblock In \emph{Proceedings of the 58th Annual Meeting of the Association for Computational Linguistics, {ACL} 2020, Online, July 5-10, 2020}, pages 7871--7880. Association for Computational Linguistics.

\bibitem[{Liang et~al.(2023)Liang, He, and Tan}]{da_li23}
Jian Liang, Ran He, and Tieniu Tan. 2023.
\newblock \href {https://doi.org/10.48550/ARXIV.2303.15361} {A comprehensive survey on test-time adaptation under distribution shifts}.
\newblock \emph{CoRR}, abs/2303.15361.

\bibitem[{Liu et~al.(2019)Liu, Ott, Goyal, Du, Joshi, Chen, Levy, Lewis, Zettlemoyer, and Stoyanov}]{roberta}
Yinhan Liu, Myle Ott, Naman Goyal, Jingfei Du, Mandar Joshi, Danqi Chen, Omer Levy, Mike Lewis, Luke Zettlemoyer, and Veselin Stoyanov. 2019.
\newblock \href {http://arxiv.org/abs/1907.11692} {Roberta: {A} robustly optimized {BERT} pretraining approach}.
\newblock \emph{CoRR}, abs/1907.11692.

\bibitem[{Lv et~al.(2023)Lv, Zhang, Shen, and Corporation}]{neural_chat}
Kaokao Lv, Wenxin Zhang, Haihao Shen, and Intel Corporation. 2023.
\newblock \href {https://medium.com/intel-analytics-software/the-practice-of-supervised-finetuning-and-direct-preference-optimization-on-habana-gaudi2-a1197d8a3cd3} {Supervised fine-tuning and direct preference optimization on intel gaudi2}.

\bibitem[{Magar and Schwartz(2022)}]{contamination}
Inbal Magar and Roy Schwartz. 2022.
\newblock \href {https://doi.org/10.18653/V1/2022.ACL-SHORT.18} {Data contamination: From memorization to exploitation}.
\newblock In \emph{Proceedings of the 60th Annual Meeting of the Association for Computational Linguistics (Volume 2: Short Papers), {ACL} 2022, Dublin, Ireland, May 22-27, 2022}, pages 157--165. Association for Computational Linguistics.

\bibitem[{McCoy et~al.(2019)McCoy, Pavlick, and Linzen}]{nli_hans}
Tom McCoy, Ellie Pavlick, and Tal Linzen. 2019.
\newblock \href {https://doi.org/10.18653/v1/p19-1334} {Right for the wrong reasons: Diagnosing syntactic heuristics in natural language inference}.
\newblock In \emph{Proceedings of the 57th Conference of the Association for Computational Linguistics, {ACL} 2019, Florence, Italy, July 28- August 2, 2019, Volume 1: Long Papers}, pages 3428--3448. Association for Computational Linguistics.

\bibitem[{Miller et~al.(2020)Miller, Krauth, Recht, and Schmidt}]{robustness_qa}
John Miller, Karl Krauth, Benjamin Recht, and Ludwig Schmidt. 2020.
\newblock \href {http://proceedings.mlr.press/v119/miller20a.html} {The effect of natural distribution shift on question answering models}.
\newblock In \emph{Proceedings of the 37th International Conference on Machine Learning, {ICML} 2020, 13-18 July 2020, Virtual Event}, volume 119 of \emph{Proceedings of Machine Learning Research}, pages 6905--6916. {PMLR}.

\bibitem[{Miller et~al.(2021)Miller, Laparra, and Bethard}]{dg_example_health_ner}
Timothy Miller, Egoitz Laparra, and Steven Bethard. 2021.
\newblock \href {https://aclanthology.org/2021.adaptnlp-1.11} {Domain adaptation in practice: Lessons from a real-world information extraction pipeline}.
\newblock In \emph{Proceedings of the Second Workshop on Domain Adaptation for NLP}, pages 105--110, Kyiv, Ukraine. Association for Computational Linguistics.

\bibitem[{Min et~al.(2022)Min, Lyu, Holtzman, Artetxe, Lewis, Hajishirzi, and Zettlemoyer}]{prompt_sensitivity}
Sewon Min, Xinxi Lyu, Ari Holtzman, Mikel Artetxe, Mike Lewis, Hannaneh Hajishirzi, and Luke Zettlemoyer. 2022.
\newblock \href {https://aclanthology.org/2022.emnlp-main.759} {Rethinking the role of demonstrations: What makes in-context learning work?}
\newblock In \emph{Proceedings of the 2022 Conference on Empirical Methods in Natural Language Processing, {EMNLP} 2022, Abu Dhabi, United Arab Emirates, December 7-11, 2022}, pages 11048--11064. Association for Computational Linguistics.

\bibitem[{Mitra et~al.(2023)Mitra, Corro, Mahajan, Codas, Sim{\~{o}}es, Agrawal, Chen, Razdaibiedina, Jones, Aggarwal, Palangi, Zheng, Rosset, Khanpour, and Awadallah}]{orca_v2}
Arindam Mitra, Luciano~Del Corro, Shweti Mahajan, Andr{\'{e}}s Codas, Clarisse Sim{\~{o}}es, Sahaj Agrawal, Xuxi Chen, Anastasia Razdaibiedina, Erik Jones, Kriti Aggarwal, Hamid Palangi, Guoqing Zheng, Corby Rosset, Hamed Khanpour, and Ahmed Awadallah. 2023.
\newblock \href {https://doi.org/10.48550/ARXIV.2311.11045} {Orca 2: Teaching small language models how to reason}.
\newblock \emph{CoRR}, abs/2311.11045.

\bibitem[{Mosbach et~al.(2023)Mosbach, Pimentel, Ravfogel, Klakow, and Elazar}]{ft_vs_fs}
Marius Mosbach, Tiago Pimentel, Shauli Ravfogel, Dietrich Klakow, and Yanai Elazar. 2023.
\newblock \href {https://doi.org/10.18653/V1/2023.FINDINGS-ACL.779} {Few-shot fine-tuning vs. in-context learning: {A} fair comparison and evaluation}.
\newblock In \emph{Findings of the Association for Computational Linguistics: {ACL} 2023, Toronto, Canada, July 9-14, 2023}, pages 12284--12314. Association for Computational Linguistics.

\bibitem[{Mukherjee et~al.(2023)Mukherjee, Mitra, Jawahar, Agarwal, Palangi, and Awadallah}]{orca_dataset}
Subhabrata Mukherjee, Arindam Mitra, Ganesh Jawahar, Sahaj Agarwal, Hamid Palangi, and Ahmed Awadallah. 2023.
\newblock \href {https://doi.org/10.48550/ARXIV.2306.02707} {Orca: Progressive learning from complex explanation traces of {GPT-4}}.
\newblock \emph{CoRR}, abs/2306.02707.

\bibitem[{Nguyen(2015)}]{airline}
Quang Nguyen. 2015.
\newblock \href {https://github.com/quankiquanki/skytrax-reviews-dataset} {The airline review dataset}.

\bibitem[{OpenAI(2023)}]{gpt4}
OpenAI. 2023.
\newblock \href {https://doi.org/10.48550/ARXIV.2303.08774} {{GPT-4} technical report}.
\newblock \emph{CoRR}, abs/2303.08774.

\bibitem[{Ormazabal et~al.(2023)Ormazabal, Artetxe, and Agirre}]{comblm}
Aitor Ormazabal, Mikel Artetxe, and Eneko Agirre. 2023.
\newblock \href {https://aclanthology.org/2023.emnlp-main.180} {Comblm: Adapting black-box language models through small fine-tuned models}.
\newblock In \emph{Proceedings of the 2023 Conference on Empirical Methods in Natural Language Processing, {EMNLP} 2023, Singapore, December 6-10, 2023}, pages 2961--2974. Association for Computational Linguistics.

\bibitem[{Plank and van Noord(2011)}]{da_data_selection}
Barbara Plank and Gertjan van Noord. 2011.
\newblock \href {https://aclanthology.org/P11-1157/} {Effective measures of domain similarity for parsing}.
\newblock In \emph{The 49th Annual Meeting of the Association for Computational Linguistics: Human Language Technologies, Proceedings of the Conference, 19-24 June, 2011, Portland, Oregon, {USA}}, pages 1566--1576. The Association for Computer Linguistics.

\bibitem[{Pontiki et~al.(2016)Pontiki, Galanis, Papageorgiou, Androutsopoulos, and Suresh~Manandhar}]{semeval_2016_absa}
Maria Pontiki, Dimitris Galanis, Haris Papageorgiou, Ion Androutsopoulos, and et~al. Suresh~Manandhar. 2016.
\newblock \href {https://doi.org/10.18653/v1/s16-1002} {Semeval-2016 task 5: Aspect based sentiment analysis}.
\newblock In \emph{Proceedings of the 10th International Workshop on Semantic Evaluation, SemEval@NAACL-HLT 2016, San Diego, CA, USA, June 16-17, 2016}, pages 19--30. The Association for Computer Linguistics.

\bibitem[{Pontiki et~al.(2015)Pontiki, Galanis, Papageorgiou, Manandhar, and Androutsopoulos}]{semeval_2015_absa}
Maria Pontiki, Dimitris Galanis, Haris Papageorgiou, Suresh Manandhar, and Ion Androutsopoulos. 2015.
\newblock \href {https://doi.org/10.18653/v1/s15-2082} {Semeval-2015 task 12: Aspect based sentiment analysis}.
\newblock In \emph{Proceedings of the 9th International Workshop on Semantic Evaluation, SemEval@NAACL-HLT 2015, Denver, Colorado, USA, June 4-5, 2015}, pages 486--495. The Association for Computer Linguistics.

\bibitem[{Pontiki et~al.(2014)Pontiki, Galanis, Pavlopoulos, Papageorgiou, Androutsopoulos, and Manandhar}]{semeval_2014_absa}
Maria Pontiki, Dimitris Galanis, John Pavlopoulos, Harris Papageorgiou, Ion Androutsopoulos, and Suresh Manandhar. 2014.
\newblock \href {https://doi.org/10.3115/v1/s14-2004} {Semeval-2014 task 4: Aspect based sentiment analysis}.
\newblock In \emph{Proceedings of the 8th International Workshop on Semantic Evaluation, SemEval@COLING 2014, Dublin, Ireland, August 23-24, 2014}, pages 27--35. The Association for Computer Linguistics.

\bibitem[{Radford et~al.(2019)Radford, Wu, Child, Luan, Amodei, Sutskever et~al.}]{gpt2}
Alec Radford, Jeffrey Wu, Rewon Child, David Luan, Dario Amodei, Ilya Sutskever, et~al. 2019.
\newblock Language models are unsupervised multitask learners.
\newblock \emph{OpenAI blog}, 1(8):9.

\bibitem[{Raffel et~al.(2020)Raffel, Shazeer, Roberts, Lee, Narang, Matena, Zhou, Li, and Liu}]{t5}
Colin Raffel, Noam Shazeer, Adam Roberts, Katherine Lee, Sharan Narang, Michael Matena, Yanqi Zhou, Wei Li, and Peter~J. Liu. 2020.
\newblock \href {http://jmlr.org/papers/v21/20-074.html} {Exploring the limits of transfer learning with a unified text-to-text transformer}.
\newblock \emph{J. Mach. Learn. Res.}, 21:140:1--140:67.

\bibitem[{Rajpurkar et~al.(2018)Rajpurkar, Jia, and Liang}]{squad_v2}
Pranav Rajpurkar, Robin Jia, and Percy Liang. 2018.
\newblock \href {https://doi.org/10.18653/v1/P18-2124} {Know what you don't know: Unanswerable questions for squad}.
\newblock In \emph{Proceedings of the 56th Annual Meeting of the Association for Computational Linguistics, {ACL} 2018, Melbourne, Australia, July 15-20, 2018, Volume 2: Short Papers}, pages 784--789. Association for Computational Linguistics.

\bibitem[{Rajpurkar et~al.(2016)Rajpurkar, Zhang, Lopyrev, and Liang}]{squad_v1}
Pranav Rajpurkar, Jian Zhang, Konstantin Lopyrev, and Percy Liang. 2016.
\newblock \href {https://doi.org/10.18653/v1/d16-1264} {Squad: 100, 000+ questions for machine comprehension of text}.
\newblock In \emph{Proceedings of the 2016 Conference on Empirical Methods in Natural Language Processing, {EMNLP} 2016, Austin, Texas, USA, November 1-4, 2016}, pages 2383--2392. The Association for Computational Linguistics.

\bibitem[{Ramponi and Plank(2020)}]{unsupervised_da_survey}
Alan Ramponi and Barbara Plank. 2020.
\newblock \href {https://doi.org/10.18653/v1/2020.coling-main.603} {Neural unsupervised domain adaptation in {NLP} - {A} survey}.
\newblock In \emph{Proceedings of the 28th International Conference on Computational Linguistics, {COLING} 2020, Barcelona, Spain (Online), December 8-13, 2020}, pages 6838--6855. International Committee on Computational Linguistics.

\bibitem[{Redko et~al.(2020)Redko, Morvant, Habrard, Sebban, and Bennani}]{divergence_theory}
Ievgen Redko, Emilie Morvant, Amaury Habrard, Marc Sebban, and Youn{\`{e}}s Bennani. 2020.
\newblock \href {http://arxiv.org/abs/2004.11829} {A survey on domain adaptation theory}.
\newblock \emph{CoRR}, abs/2004.11829.

\bibitem[{Reid et~al.(2022)Reid, Zhong, Gururangan, and Zettlemoyer}]{lm_da_benchmark}
Machel Reid, Victor Zhong, Suchin Gururangan, and Luke Zettlemoyer. 2022.
\newblock \href {https://aclanthology.org/2022.emnlp-main.63} {{M2D2:} {A} massively multi-domain language modeling dataset}.
\newblock In \emph{Proceedings of the 2022 Conference on Empirical Methods in Natural Language Processing, {EMNLP} 2022, Abu Dhabi, United Arab Emirates, December 7-11, 2022}, pages 964--975. Association for Computational Linguistics.

\bibitem[{Reimers and Gurevych(2019)}]{sentence_transformer}
Nils Reimers and Iryna Gurevych. 2019.
\newblock \href {https://doi.org/10.18653/v1/D19-1410} {Sentence-bert: Sentence embeddings using siamese bert-networks}.
\newblock In \emph{Proceedings of the 2019 Conference on Empirical Methods in Natural Language Processing and the 9th International Joint Conference on Natural Language Processing, {EMNLP-IJCNLP} 2019, Hong Kong, China, November 3-7, 2019}, pages 3980--3990. Association for Computational Linguistics.

\bibitem[{Remus(2012)}]{remus_divergence}
Robert Remus. 2012.
\newblock \href {https://doi.org/10.1109/ICDMW.2012.46} {Domain adaptation using domain similarity- and domain complexity-based instance selection for cross-domain sentiment analysis}.
\newblock In \emph{12th {IEEE} International Conference on Data Mining Workshops, {ICDM} Workshops, Brussels, Belgium, December 10, 2012}, pages 717--723. {IEEE} Computer Society.

\bibitem[{Rotman and Reichart(2019)}]{self_training_rotman}
Guy Rotman and Roi Reichart. 2019.
\newblock \href {https://transacl.org/ojs/index.php/tacl/article/view/1801} {Deep contextualized self-training for low resource dependency parsing}.
\newblock \emph{Trans. Assoc. Comput. Linguistics}, 7:695--713.

\bibitem[{Ruder et~al.(2017)Ruder, Ghaffari, and Breslin}]{ruder_divergence}
Sebastian Ruder, Parsa Ghaffari, and John~G. Breslin. 2017.
\newblock \href {http://arxiv.org/abs/1702.02426} {Data selection strategies for multi-domain sentiment analysis}.
\newblock \emph{CoRR}, abs/1702.02426.

\bibitem[{Rychalska et~al.(2019)Rychalska, Basaj, Gosiewska, and Biecek}]{wildnlp}
Barbara Rychalska, Dominika Basaj, Alicja Gosiewska, and Przemyslaw Biecek. 2019.
\newblock \href {https://doi.org/10.1007/978-3-030-36718-3\_20} {Models in the wild: On corruption robustness of neural {NLP} systems}.
\newblock In \emph{Neural Information Processing - 26th International Conference, {ICONIP} 2019, Sydney, NSW, Australia, December 12-15, 2019, Proceedings, Part {III}}, volume 11955 of \emph{Lecture Notes in Computer Science}, pages 235--247. Springer.

\bibitem[{Sanh et~al.(2019)Sanh, Debut, Chaumond, and Wolf}]{distilbert}
Victor Sanh, Lysandre Debut, Julien Chaumond, and Thomas Wolf. 2019.
\newblock \href {http://arxiv.org/abs/1910.01108} {Distilbert, a distilled version of {BERT:} smaller, faster, cheaper and lighter}.
\newblock \emph{CoRR}, abs/1910.01108.

\bibitem[{Shi et~al.(2023)Shi, Ajith, Xia, Huang, Liu, Blevins, Chen, and Zettlemoyer}]{contamination2}
Weijia Shi, Anirudh Ajith, Mengzhou Xia, Yangsibo Huang, Daogao Liu, Terra Blevins, Danqi Chen, and Luke Zettlemoyer. 2023.
\newblock \href {https://doi.org/10.48550/ARXIV.2310.16789} {Detecting pretraining data from large language models}.
\newblock \emph{CoRR}, abs/2310.16789.

\bibitem[{Touvron et~al.(2023)Touvron, Martin, Stone, and Peter~Albert}]{llama_v2}
Hugo Touvron, Louis Martin, Kevin Stone, and et.~al. Peter~Albert. 2023.
\newblock \href {https://doi.org/10.48550/ARXIV.2307.09288} {Llama 2: Open foundation and fine-tuned chat models}.
\newblock \emph{CoRR}, abs/2307.09288.

\bibitem[{Tu et~al.(2020)Tu, Lalwani, Gella, and He}]{dr_tu}
Lifu Tu, Garima Lalwani, Spandana Gella, and He~He. 2020.
\newblock \href {https://doi.org/10.1162/TACL\_A\_00335} {An empirical study on robustness to spurious correlations using pre-trained language models}.
\newblock \emph{Trans. Assoc. Comput. Linguistics}, 8:621--633.

\bibitem[{Van~Asch and Daelemans(2010)}]{predict_dr2}
Vincent Van~Asch and Walter Daelemans. 2010.
\newblock \href {https://aclanthology.org/W10-2605} {Using domain similarity for performance estimation}.
\newblock In \emph{Proceedings of the 2010 Workshop on Domain Adaptation for Natural Language Processing}, pages 31--36, Uppsala, Sweden. Association for Computational Linguistics.

\bibitem[{Varshney et~al.(2022)Varshney, Mishra, and Baral}]{predict_no_asnwer}
Neeraj Varshney, Swaroop Mishra, and Chitta Baral. 2022.
\newblock \href {https://doi.org/10.18653/v1/2022.repl4nlp-1.23} {Towards improving selective prediction ability of {NLP} systems}.
\newblock In \emph{Proceedings of the 7th Workshop on Representation Learning for NLP, RepL4NLP@ACL 2022, Dublin, Ireland, May 26, 2022}, pages 221--226. Association for Computational Linguistics.

\bibitem[{Veen et~al.(2023)Veen, Uden, Attias, Pareek, Bluethgen, Polacin, Chiu, Delbrouck, Chaves, Langlotz, Chaudhari, and Pauly}]{da_va23}
Dave~Van Veen, Cara~Van Uden, Maayane Attias, Anuj Pareek, Christian Bluethgen, Malgorzata Polacin, Wah Chiu, Jean{-}Benoit Delbrouck, Juan Manuel~Zambrano Chaves, Curtis~P. Langlotz, Akshay Chaudhari, and John~M. Pauly. 2023.
\newblock \href {https://doi.org/10.18653/V1/2023.BIONLP-1.42} {Radadapt: Radiology report summarization via lightweight domain adaptation of large language models}.
\newblock In \emph{The 22nd Workshop on Biomedical Natural Language Processing and BioNLP Shared Tasks, BioNLP@ACL 2023, Toronto, Canada, 13 July 2023}, pages 449--460. Association for Computational Linguistics.

\bibitem[{Volk et~al.(2022)Volk, Ben{-}David, Amosy, Chechik, and Reichart}]{hyperpada}
Tomer Volk, Eyal Ben{-}David, Ohad Amosy, Gal Chechik, and Roi Reichart. 2022.
\newblock \href {https://doi.org/10.48550/arXiv.2203.14276} {Example-based hypernetworks for out-of-distribution generalization}.
\newblock \emph{CoRR}, abs/2203.14276.

\bibitem[{V{\"{o}}lske et~al.(2017)V{\"{o}}lske, Potthast, Syed, and Stein}]{reddit_tldr}
Michael V{\"{o}}lske, Martin Potthast, Shahbaz Syed, and Benno Stein. 2017.
\newblock \href {https://doi.org/10.18653/v1/w17-4508} {Tl;dr: Mining reddit to learn automatic summarization}.
\newblock In \emph{Proceedings of the Workshop on New Frontiers in Summarization, NFiS@EMNLP 2017, Copenhagen, Denmark, September 7, 2017}, pages 59--63. Association for Computational Linguistics.

\bibitem[{Wang et~al.(2019)Wang, Pruksachatkun, Nangia, Singh, Michael, Hill, Levy, and Bowman}]{superglue}
Alex Wang, Yada Pruksachatkun, Nikita Nangia, Amanpreet Singh, Julian Michael, Felix Hill, Omer Levy, and Samuel~R. Bowman. 2019.
\newblock \href {https://proceedings.neurips.cc/paper/2019/hash/4496bf24afe7fab6f046bf4923da8de6-Abstract.html} {Superglue: {A} stickier benchmark for general-purpose language understanding systems}.
\newblock In \emph{Advances in Neural Information Processing Systems 32: Annual Conference on Neural Information Processing Systems 2019, NeurIPS 2019, December 8-14, 2019, Vancouver, BC, Canada}, pages 3261--3275.

\bibitem[{Wang et~al.(2022{\natexlab{a}})Wang, Roberts, Hesslow, Scao, Chung, Beltagy, Launay, and Raffel}]{zero_shot_arch}
Thomas Wang, Adam Roberts, Daniel Hesslow, Teven~Le Scao, Hyung~Won Chung, Iz~Beltagy, Julien Launay, and Colin Raffel. 2022{\natexlab{a}}.
\newblock \href {https://proceedings.mlr.press/v162/wang22u.html} {What language model architecture and pretraining objective works best for zero-shot generalization?}
\newblock In \emph{International Conference on Machine Learning, {ICML} 2022, 17-23 July 2022, Baltimore, Maryland, {USA}}, volume 162 of \emph{Proceedings of Machine Learning Research}, pages 22964--22984. {PMLR}.

\bibitem[{Wang et~al.(2022{\natexlab{b}})Wang, Wang, and Yang}]{robustness_nlp_survey}
Xuezhi Wang, Haohan Wang, and Diyi Yang. 2022{\natexlab{b}}.
\newblock \href {https://doi.org/10.18653/v1/2022.naacl-main.339} {Measure and improve robustness in {NLP} models: {A} survey}.
\newblock In \emph{Proceedings of the 2022 Conference of the North American Chapter of the Association for Computational Linguistics: Human Language Technologies, {NAACL} 2022, Seattle, WA, United States, July 10-15, 2022}, pages 4569--4586. Association for Computational Linguistics.

\bibitem[{Weber et~al.(2023)Weber, Bruni, and Hupkes}]{prompt_robustness}
Lucas Weber, Elia Bruni, and Dieuwke Hupkes. 2023.
\newblock \href {https://aclanthology.org/2023.conll-1.20} {Mind the instructions: a holistic evaluation of consistency and interactions in prompt-based learning}.
\newblock In \emph{Proceedings of the 27th Conference on Computational Natural Language Learning, CoNLL 2023, Singapore, December 6-7, 2023}, pages 294--313. Association for Computational Linguistics.

\bibitem[{Williams et~al.(2018)Williams, Nangia, and Bowman}]{mnli}
Adina Williams, Nikita Nangia, and Samuel~R. Bowman. 2018.
\newblock \href {https://doi.org/10.18653/v1/n18-1101} {A broad-coverage challenge corpus for sentence understanding through inference}.
\newblock In \emph{Proceedings of the 2018 Conference of the North American Chapter of the Association for Computational Linguistics: Human Language Technologies, {NAACL-HLT} 2018, New Orleans, Louisiana, USA, June 1-6, 2018, Volume 1 (Long Papers)}, pages 1112--1122. Association for Computational Linguistics.

\bibitem[{Yang et~al.(2023{\natexlab{a}})Yang, Liu, Li, Yin, You, Yin, and Zou}]{titles}
Bang Yang, Fenglin Liu, Zheng Li, Qingyu Yin, Chenyu You, Bing Yin, and Yuexian Zou. 2023{\natexlab{a}}.
\newblock \href {https://doi.org/10.18653/V1/2023.FINDINGS-ACL.166} {Multimodal prompt learning for product title generation with extremely limited labels}.
\newblock In \emph{Findings of the Association for Computational Linguistics: {ACL} 2023, Toronto, Canada, July 9-14, 2023}, pages 2652--2665. Association for Computational Linguistics.

\bibitem[{Yang et~al.(2023{\natexlab{b}})Yang, Song, Ren, Lyu, Wang, Zhuo, Liu, Wang, Foster, and Zhang}]{da_ya23}
Linyi Yang, Yaoxian Song, Xuan Ren, Chenyang Lyu, Yidong Wang, Jingming Zhuo, Lingqiao Liu, Jindong Wang, Jennifer Foster, and Yue Zhang. 2023{\natexlab{b}}.
\newblock \href {https://aclanthology.org/2023.emnlp-main.276} {Out-of-distribution generalization in natural language processing: Past, present, and future}.
\newblock In \emph{Proceedings of the 2023 Conference on Empirical Methods in Natural Language Processing, {EMNLP} 2023, Singapore, December 6-10, 2023}, pages 4533--4559. Association for Computational Linguistics.

\bibitem[{Yu et~al.(2023)Yu, Wang, Golovneva, AlKhamissi, Verma, Jin, Ghosh, Diab, and Celikyilmaz}]{yu-etal-2023-alert}
Ping Yu, Tianlu Wang, Olga Golovneva, Badr AlKhamissi, Siddharth Verma, Zhijing Jin, Gargi Ghosh, Mona Diab, and Asli Celikyilmaz. 2023.
\newblock \href {https://doi.org/10.18653/v1/2023.acl-long.60} {{ALERT}: Adapt language models to reasoning tasks}.
\newblock In \emph{Proceedings of the 61st Annual Meeting of the Association for Computational Linguistics (Volume 1: Long Papers)}, pages 1055--1081, Toronto, Canada. Association for Computational Linguistics.

\bibitem[{Yu et~al.(2021)Yu, Liu, and Fung}]{adaptsum_summarization_dataset}
Tiezheng Yu, Zihan Liu, and Pascale Fung. 2021.
\newblock \href {https://doi.org/10.18653/v1/2021.naacl-main.471} {Adaptsum: Towards low-resource domain adaptation for abstractive summarization}.
\newblock In \emph{Proceedings of the 2021 Conference of the North American Chapter of the Association for Computational Linguistics: Human Language Technologies, {NAACL-HLT} 2021, Online, June 6-11, 2021}, pages 5892--5904. Association for Computational Linguistics.

\bibitem[{Yu et~al.(2022)Yu, Khan, and Xu}]{measuring_robustness}
Yu~Yu, Abdul~Rafae Khan, and Jia Xu. 2022.
\newblock \href {https://aclanthology.org/2022.coling-1.343} {Measuring robustness for {NLP}}.
\newblock In \emph{Proceedings of the 29th International Conference on Computational Linguistics, {COLING} 2022, Gyeongju, Republic of Korea, October 12-17, 2022}, pages 3908--3916. International Committee on Computational Linguistics.

\bibitem[{Yuan et~al.(2023)Yuan, Chen, Cui, Gao, Zou, Cheng, Ji, Liu, and Sun}]{revisiting}
Lifan Yuan, Yangyi Chen, Ganqu Cui, Hongcheng Gao, Fangyuan Zou, Xingyi Cheng, Heng Ji, Zhiyuan Liu, and Maosong Sun. 2023.
\newblock \href {https://doi.org/10.48550/ARXIV.2306.04618} {Revisiting out-of-distribution robustness in {NLP:} benchmark, analysis, and llms evaluations}.
\newblock \emph{CoRR}, abs/2306.04618.

\bibitem[{Zhang et~al.(2020)Zhang, Kishore, Wu, Weinberger, and Artzi}]{bertscore}
Tianyi Zhang, Varsha Kishore, Felix Wu, Kilian~Q. Weinberger, and Yoav Artzi. 2020.
\newblock \href {https://openreview.net/forum?id=SkeHuCVFDr} {Bertscore: Evaluating text generation with {BERT}}.
\newblock In \emph{8th International Conference on Learning Representations, {ICLR} 2020, Addis Ababa, Ethiopia, April 26-30, 2020}. OpenReview.net.

\bibitem[{Zhong et~al.(2021)Zhong, Yin, Yu, Zaidi, Mutuma, Jha, Awadallah, Celikyilmaz, Liu, Qiu, and Radev}]{qmsum_meeting_summarization_dataset}
Ming Zhong, Da~Yin, Tao Yu, Ahmad Zaidi, Mutethia Mutuma, Rahul Jha, Ahmed~Hassan Awadallah, Asli Celikyilmaz, Yang Liu, Xipeng Qiu, and Dragomir~R. Radev. 2021.
\newblock \href {https://doi.org/10.18653/v1/2021.naacl-main.472} {Qmsum: {A} new benchmark for query-based multi-domain meeting summarization}.
\newblock In \emph{Proceedings of the 2021 Conference of the North American Chapter of the Association for Computational Linguistics: Human Language Technologies, {NAACL-HLT} 2021, Online, June 6-11, 2021}, pages 5905--5921. Association for Computational Linguistics.

\bibitem[{Ziser and Reichart(2017)}]{ziser1}
Yftah Ziser and Roi Reichart. 2017.
\newblock \href {https://doi.org/10.18653/V1/K17-1040} {Neural structural correspondence learning for domain adaptation}.
\newblock In \emph{Proceedings of the 21st Conference on Computational Natural Language Learning (CoNLL 2017), Vancouver, Canada, August 3-4, 2017}, pages 400--410. Association for Computational Linguistics.

\bibitem[{Ziser and Reichart(2018)}]{airline_ziser}
Yftah Ziser and Roi Reichart. 2018.
\newblock \href {https://doi.org/10.18653/v1/n18-1112} {Pivot based language modeling for improved neural domain adaptation}.
\newblock In \emph{Proceedings of the 2018 Conference of the North American Chapter of the Association for Computational Linguistics: Human Language Technologies, {NAACL-HLT} 2018, New Orleans, Louisiana, USA, June 1-6, 2018, Volume 1 (Long Papers)}, pages 1241--1251. Association for Computational Linguistics.

\bibitem[{Ziser and Reichart(2019)}]{ziser3}
Yftah Ziser and Roi Reichart. 2019.
\newblock \href {https://doi.org/10.18653/V1/P19-1591} {Task refinement learning for improved accuracy and stability of unsupervised domain adaptation}.
\newblock In \emph{Proceedings of the 57th Conference of the Association for Computational Linguistics, {ACL} 2019, Florence, Italy, July 28- August 2, 2019, Volume 1: Long Papers}, pages 5895--5906. Association for Computational Linguistics.

\end{thebibliography}
\clearpage

\appendix
\renewcommand \thepart{}
\renewcommand \partname{}
\mtcsettitle{parttoc}{}
\addcontentsline{toc}{section}{Appendix} 
\part{Appendix} 
\parttoc 

\newpage

\section{On The Relationship Between SS, TT, ST, SD and TD}
\label{sec:theorem}

In this subsection, we expand the discussion from \S\ref{sub:metrics} and \S\ref{sub:drops} about the Domain Robustness (DR) metrics introduced in our study. Our aim is to address and clarify any questions that might arise from the nuanced definitions presented earlier. Additionally, we offer a theoretical perspective on our findings discussed in the results subsection \S\ref{sub:comparing}.

In \S\ref{sec:methods} we define the DR challenge as \textit{the inherent inability of an NLP model to generalize from the source domain to the target domains.} This inability is closely linked to the in-domain and cross-domain performance of the model, and \textit{full characterization of it requires understanding the joint distribution of $\sss$, $\ttt$ and $\st$.} These three \textit{performance measures} are random variables, with their variability stemming from the selection of source and target domains, the sampling of training and testing data from these domains, and the variabilities in the training and inference processes. 

Nonetheless, identifying these random variables and their relationships is not tractable without further assumptions. We hence introduce simple, practical, and interpretable metrics that quantify the properties of the joint distribution: 
$\overline{\sss} = \E[\sss]$, $\overline{\st} = \E[\st]$ and $\drop = \overline{\sss} - \overline{\st}$. Although our definitions rely on expectations, in practice, these metrics are task-level statistics (averages) that estimate them. Intuitively, the average drop ($\drop$) estimates the expected task-level performance degradation when shifting domains and \textit{the larger it is, the more severe the DR challenge of the model is.} 

Other three metrics that are derived from the joint distribution and quantify performance degradation at the shift level are $\sd$, $\td$, and $\idd$. A positive in-domain difference may indicate a shift to a \textit{harder domain}, and in contrast to the $\sd$ and the $\td$, \textit{the $\idd$ is not a genuine by-product of the DR challenge since it does not consider the $\st$.} 

Based on our assertion that the joint distribution of $\sss$, $\ttt$ and $\st$ is needed for characterizing the DR challenge, then it follows that we need at least two of the \textit{degradation metrics} ($\sd$, $\td$, and $\idd$) to do so. Moreover, the following trivial equation:
\begin{align*}
\sd = \idd + \td \\
\td = \idd - \sd
\end{align*}
\noindent presents how the three metrics are connected. Accordingly, looking solely on one drop metric ($\sd$ or $\td$) can lead to incorrect conclusions, as large drops might be attributed to the $\idd$. Notably, when a range of experiments is conducted \textbf{using all domains for both training and testing}, it follows that $\E[\sss] = \E[\ttt]$, and from the linearity of the expectation, $\drop = \E[\sd] = \E[\td]$. Importantly, while $\sd$ and $\td$ have equal expected values, they are distinct random variables with differing variances. See the toy example in Table~\ref{tab:toy}.

Although an accurate and truthful understanding of the DR challenge requires considering both metrics, many works measure only the $\sd$. However, this is the least indicative option, as we empirically show that the $\td$ is a more robust estimator of the average drop, $\drop$. This is because the $\td$ tends to have a lower extreme magnitude and variance than the $\sd$, and the $\idd$ explains a larger portion of the $\sd$ than the $\td$.

Below, we introduce a theorem that binds these properties together and demonstrates their equivalence. But even more, it reveals them to be equivalent to the case when the $\st$ is more akin to the $\ttt$ (e.g. when ${\Cov[\st, \ttt]>\Cov[\st, \sss]}$). In other words, if we believe that in our task the potential of the model to perform well cross-domain is determined by the difficulty of the target domain, as in the case of challenge sets, then the reference point for measuring a degradation should be the $\ttt$ and not the $\sss$, and the $\td$ would be indeed the better drop metric.

\begin{table}[t]
\centering
\begin{adjustbox}{width=0.48\textwidth}
\begin{tabular}{cc|cccccc|l}
\toprule
{\footnotesize \textbf{Source}} & {\footnotesize \textbf{Target}} & {\footnotesize $\st$} & {\footnotesize $\sss$} & {\footnotesize $\ttt$} & {\footnotesize $\idd$} & {\footnotesize $\sd$} & {\footnotesize $\td$} & {\footnotesize \textbf{Scenario}} \\
\midrule
A & A & & 90 & 90 & & & & \\
B & B & & 80 & 80 & & & & \\
C & C & & 70 & 70 & & & & \\
\hline
 \rowcolor{lightgray!50} A & B & 75 & 90 & 80 & 10 & 15 & 5 & Classic \\
 \rowcolor{lightgray!50} A & C & 75 & 90 & 70 & 20 & 15 & -5 & Observed \\
B & A & 95 & 80 & 90 & -10 & -15 & -5 & No Challenge \\
 \rowcolor{lightgray!50} B & C & 65 & 80 & 70 & 10 & 15 & 5 & Classic \\
C & A & 80 & 70 & 90 & -20 & -10 & 10 & Unobserved \\
C & B & 75 & 70 & 80 & -10 & -5 & 5 & Unobserved \\
\bottomrule
\end{tabular}
\end{adjustbox}
\caption{Toy example of domain shifts. The task-level statistics are: $\overline{\sss} = 80$;   $\overline{\st} = 77.5$; $\drop = 2.5$; $W_{\sd} = 15$; $W_{\td} = 10$. Notice that the mean of $\sd$ is $2.5$, equal to that of $\td$ and $\drop$. However, as many previous studies have done, examining only the challenging shifts (with $\idd > 0$, indicated by gray rows) and focusing on $\sd$ alone can obscure the real DR state. In these shifts, the mean $\sd$ is $15$, which might be misconstrued due to large $\idd$. Incorporating the $\td$ into the analysis can rectify this and avoid misinterpretations. Nonetheless, the most comprehensive approach to understanding task-level behavior is to consider all domains both as sources and targets, as we do. In this case, the means of all drops are equal: $\drop = \E[\sd] = \E[\td]$.}
\label{tab:toy}
\end{table}

\begin{theorem}
\label{theorem:eq}
Let $(S,T)$ be different source and target domains sampled independently from the domain space, and let $\triplet$ be RVs of their performances. The following are equivalent:
\begin{enumerate}[label=(\arabic*)]
    \item $\Cov[\ttt, \st]>\Cov[\sss, \st]$
    \item $\Cov[\idd, \sd]^2>\Cov[\idd, \td]^2$
    \item $\Var[\sd]>\Var[\td]$
    \item $\E[\lvert \sd \rvert]>\E[\lvert \td \rvert]$
\end{enumerate}
\end{theorem}

\begin{remark}
Although in Theorem~\ref{theorem:eq} we employ fundamental probability concepts such as expectation, variance, and covariance, our results utilize well-established and easily interpretable statistics: (1) We use the Pearson's correlation between the $\st$ and the $\sss$ or the $\ttt$; (2) We use the R-squared ($R^2$) between the $\idd$ and the $\sd$ or the $\td$. Notably, the R-squared indicates the proportion of the variability in a dependent variable ($\sd$) that is explained by the independent variable ($\idd$), serving as a gauge of the goodness of fit. We use Peasron's correlation to understand the relationship of $\sss$, $\ttt$, and $\st$ because it considers the directionality of the relationship, indicated by the sign. In contrast, here we use the $R^2$ since it focuses on the degree, ignoring the sign; (3) We use the sample standard deviation of the drops; (4) We use the maximum drops (Worst $\sd$ or $\td$); While the concepts in Theorem~\ref{theorem:eq} are not direct equivalents of these statistics, they are closely related and help elucidate our findings.
\end{remark}

\begin{remark}
Notice that, $\st = \ttt - \td$ and $\sd = \idd + \td$. Although we found a strong relationship between the $\st$ and the $\ttt$ (e.g., $\rho=0.95$ in the fine-tuning QG task) or between the $\sd$ and the $\idd$ (e.g., $R^2=0.96$ in fine-tuning QA task), this does not imply that the $\td$ is zero and no DR challenge exist. These strong correlations or high $R^2$ values merely reflect the $\td$ has a low variability. Its magnitude cannot be inferred from the correlation or $R^2$ alone. 
\end{remark}

\begin{proof}
We start be denoting $x=\Var[\sss]>0$ and ${y=\Cov[\ttt, \st]-\Cov[\sss, \st]}$. Notice that ${\E[\sss]=\E[\ttt]}$ and ${\Var[\sss]=\Var[\ttt]}$. 
\noindent From the linearity of expectation, we get:
\begin{align*}
    \E[\sd] & = \E[\sss] - \E[\st] \\
    & = \E[\ttt] - \E[\st] = \E[\td]
\end{align*}

\noindent\underline{(1) $\Leftrightarrow$ (2):} Since $S$ and $T$ are independent then ${\Cov[\sss,\ttt]=0}$. From the bilinearity of the covariance, we get:
\begin{align*}
    \Cov[\idd, \sd] = \Var[\sss] + \Cov[\sss, \ttt] \\ -\Cov[\sss, \st] + \Cov[\ttt, \st] = x+y
\end{align*}
Similarly, $\Cov[\idd, \sd]=-x+y$.

If (1) holds, then ${y>0}$. Since $x$ and $y$ are both positive, then ${(x+y)^2>(-x+y)^2}$ and (2) holds. The same is true for the other direction: if (2) holds, then $y$ must be positive, and (1) holds.

\noindent\underline{(1) $\Leftrightarrow$ (3):} From the variance of a sum, we get:
\begin{align*}
    & \Var[\sd] = \Var[\sss] -2\Cov[\st,\sss] + \Var[\st] \\
    & \Var[\td] = \Var[\ttt] -2\Cov[\st,\ttt] + \Var[\st]
\end{align*}
If (1) holds, then:
\begin{align*}
    & \Var[\sd] -  \Var[\td] = \\
    & 2(\Cov[\st,\ttt] -\Cov[\st,\sss]) > 0
\end{align*}
and (3) holds. Invert the order to prove $(3) \Rightarrow (1)$.

\noindent\underline{(1) $\Leftrightarrow$ (4):} Notice that:
\begin{align*}
    & \E[\sd^2] = \E[\sss]^2-2\E[\sss \cdot \st] + \E[\st]^2 \\
    & \E[\td^2] = \E[\sss]^2-2\E[\ttt \cdot \st] + \E[\st]^2
\end{align*}
Since ${\E[\sss]=\E[\ttt]}$, we get:
\begin{align*}
    \E[\sd^2] - \E[\td^2] = 2(\E[\ttt \cdot \st] - \E[\sss \cdot \st])
\end{align*}
From the definition of covariance: 
\begin{align*}
   & \Cov[\st,\sss]=\E[\sss \cdot \st] - \E[\sss]\E[\st] \\
   & \Cov[\st,\ttt]=\E[\ttt \cdot \st] - \E[\ttt]\E[\st] \\   
\end{align*}
and therefore: 
\begin{align*}
   & \Cov[\st,\ttt] - \Cov[\st,\sss] \\
   & = \E[\ttt \cdot \st] - \E[\sss \cdot \st]
\end{align*}
Now, if (1) holds, then $\E[\ttt \cdot \st] > \E[\sss \cdot \st]$ and $\E[\sd^2] > \E[\sd^2]$, and 
(4) holds. To prove the converse, reverse the implications.
\end{proof}

\subsection{Intuition for Domain Shift Scenarios}
\label{sub:intuition}

In \S\ref{sub:scenarios} we introduce a framework for classifying types of domain shifts into four scenariosL \textit{Classic} ($\sd > 0$ and $\td > 0$), \textit{Observed} ($\sd > 0$ but $\td < 0$), \textit{Unobserved} ($\sd < 0$ but $\td > 0$), and \textit{No Challenge} ($\sd < 0$ and $\td < 0$).

While performance degradation with respect to $\ttt$ (positive $\td$) seems intuitive (as we do not expect the model to perform better than it would have had it been trained on data from the target domain), one may wonder about the cases where $\td$ is negative. Specifically the \textit{Observed} and \textit{No Challenge} scenarios which can be counter-intuitive. 

In what follows, we will elaborate on these scenarios. First, notice that every scenario can occur if the effect of the domain shift is noisy. Second, consider the following motivation:

\textit{The No Challenge scenario} ($\sss > \st$ and $\ttt > \st$): Imagine a model trained on advanced math problems (graduate level) being applied to basic math problems (elementary level). In this case, we anticipate a \textit{No Challenge} scenario due to the simplicity of elementary problems compared to graduate-level problems ($\sss > \st$) and the model's capability to understand complex graduate-level content, which implies it can certainly handle elementary-level problems ($\ttt > \st$).

\textit{The Observed scenario} ($\sss> \st > \ttt $): Now consider the opposite direction. The model is trained on elementary math problems and applied to graduate-level problems. Obviously, we anticipate $\sss$ to be larger than $\st$. In addition, within the set of graduate-level problems, there are some introductory or ``warmup'' problems (that the model trained on the elementary-level problem can solve). Despite the presence of simpler problems within the graduate-level set, the overall complexity of this domain can prevent the model from learning even the elementary concepts when trained on graduate-level problems, and thus, $\st > \ttt$.

Notice that, indeed, the \textit{Observed} and \textit{No challenge} scenarios are the least common scenarios (see Figure~\ref{fig:scenarios}). They occur mostly in the few-shot setups and can be attributed to the weaker effect of the domain shift on few-shot models. In addition, they also occur for FT models in the QA and QG tasks where the shift effect is also weak (see Table~\ref{tab:main_fs_results}).

\begin{figure*}[!htb]
    \centering
    \includegraphics[width=\textwidth]{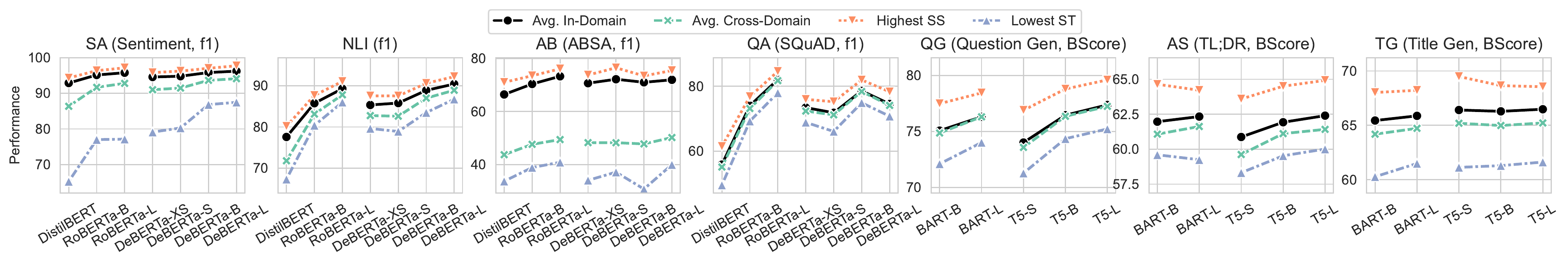}
    \caption{Fine-tuning performance for the seven tasks of different models with varying sizes. The plots present the F1 and BertScore scores of the average in-domain (black line) and cross-domain (green line) performance. In addition, the highest in-domain score (orange line) and the lowest cross-domain score (blue line) are displayed.}
    \label{fig:ft_performance}
\end{figure*}
\begin{figure*}[!htb]
    \centering
    \includegraphics[width=\textwidth]{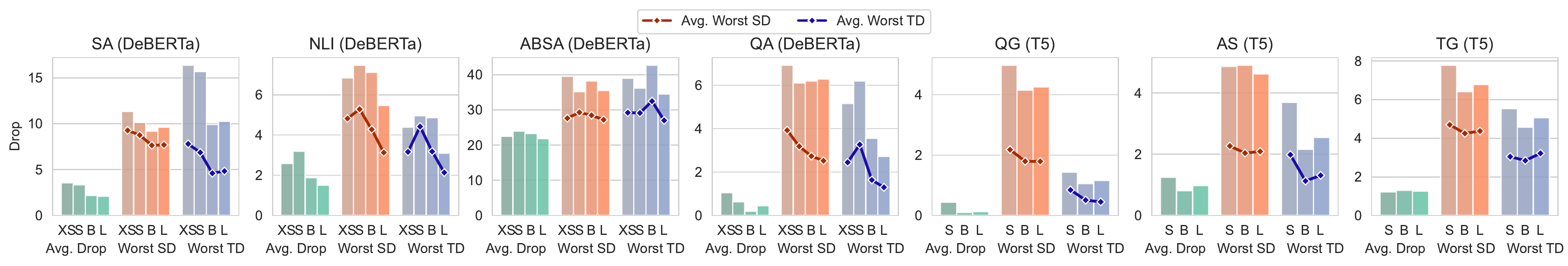}
    \caption{Fine-tuning drops of DeBERTa and T5 families. The plots present: The Average Drop (green bars); The Worst $\sd$ (orange bars); and the Worst $\td$ (blue bars). The lines on the bars present the Average Worst $\sd$ and $\td$, i.e., for each source domain we first find the worst drop and then take the average over all source domains.}
    \label{fig:ft_drops}
\end{figure*}

\section{Additional Results}
\label{sec:additional_results}

\subsection{Fine-tuned Model Size}
\label{sub:model_size}

Larger fine-tuned models often lead to better performance, but the question remains: How does the model size affect its domain robustness? To address this question, we have conducted comprehensive experiments using models of different sizes within the same architectural families, as detailed in Table~\ref{tab:models}. In Figure~\ref{fig:ft_performance}, we compare the absolute performance of various model sizes within the same model families. Conversely, Figure~\ref{fig:ft_drops} presents the performance drops for these models.

Same as our finding in \ref{sec:results}, we observe that also across all model sizes and all tasks (except QA and QG), the average in-domain performance consistently exceeds the average cross-domain performance and the Worst $\sd$ surpasses the Worst $\td$. 

When examining the influence of increasing model size, we find that, as expected, larger models within the same architectural family improve the absolute in-domain cross-domain performance. Regarding the performance drop, the general trend is that larger models reduce performance drops, a trend that is more pronounced in classification tasks. This indicates that utilizing larger models could enhance not just the absolute performance, but also the DR of these models.

\begin{figure*}[!htb]
    \centering
    \includegraphics[width=\textwidth]{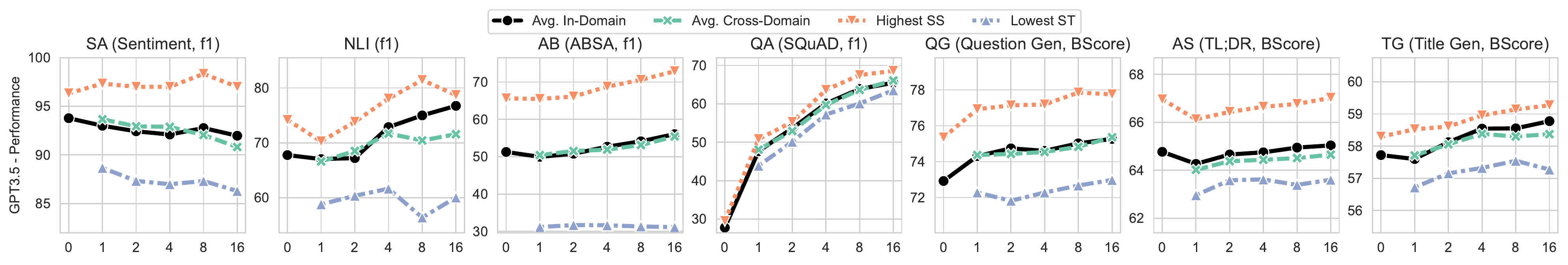}
    \includegraphics[width=\textwidth]{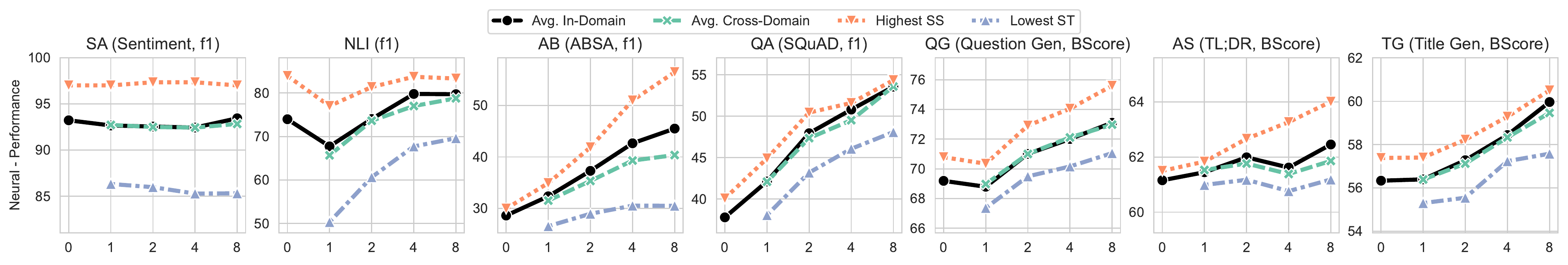}
    \includegraphics[width=\textwidth]{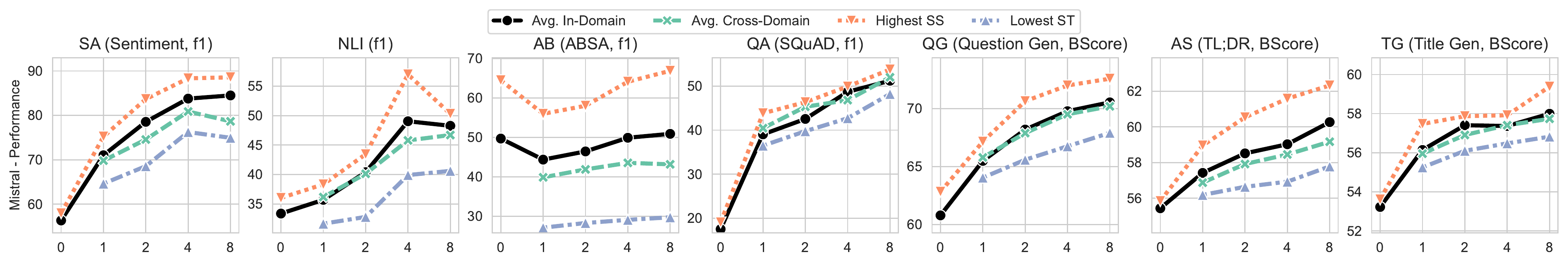}
    \caption{Performance of GPT3.5 (top), NeuralChat (middle) and Mistral (bottom) as a function of the number of few-shot demonstrations. The plots present the F1 and BertScore scores of the average in-domain (black line) and cross-domain (green line) performance. In addition, the highest in-domain score (orange line) and the lowest cross-domain score (blue line) are displayed.}
    \label{fig:demons_gpt_performance}
\end{figure*}
\begin{figure*}[!htb]
    \centering
    \includegraphics[width=\textwidth]{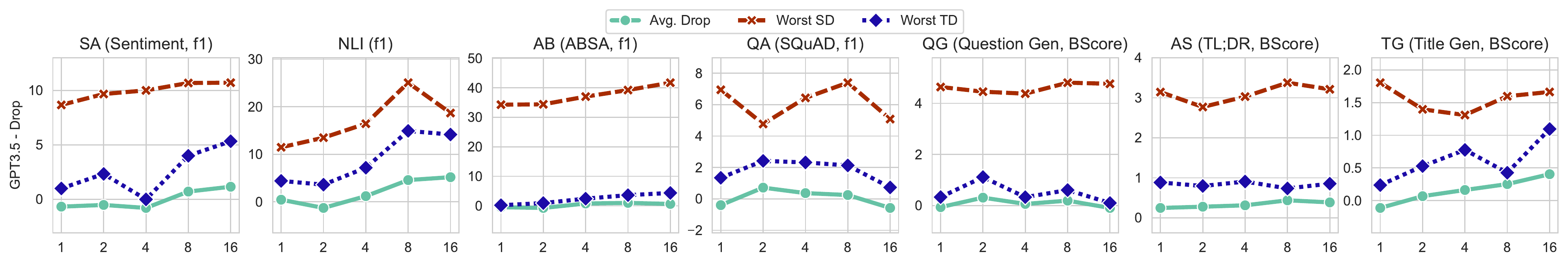}
    \includegraphics[width=\textwidth]{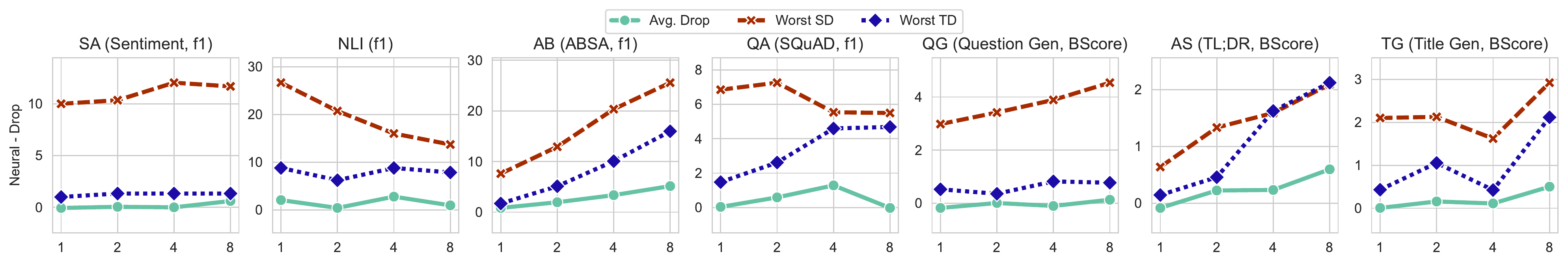}
    \includegraphics[width=\textwidth]{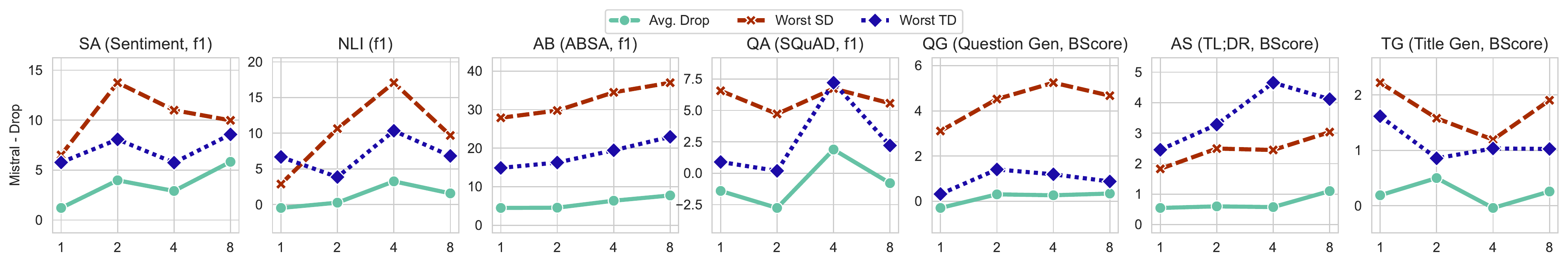}
    \caption{Performance drops of GPT3.5 (top), NeuralChat (middle) and Mistral (bottom) as a function of the number of few-shot demonstrations. The plots present: The Average Drop (green line); The Worst $\sd$ (orange line); and the Worst $\td$ (blue line).}
    \label{fig:demons_gpt_drops}
\end{figure*}

\subsection{Number of Few-shot Demonstrations}
\label{sub:number_demonstrations}

In contrast to fine-tuning, in few-shot setups there is potentially a weaker anchoring of the model in the source domain since it is not trained on domain-specific data. Instead, the few-shot model is simply provided with a few demonstrations from the source domain. We investigate whether increasing the number of demonstrations strengthens this anchoring, thereby potentially affecting the model's domain robustness. Figures~\ref{fig:demons_gpt_performance} and \ref{fig:demons_gpt_drops} illustrate the impact of the number of demonstrations on both the absolute performance and performance drops of few-shot models, respectively.

Unsurprisingly, when comparing zero-shot to few-shot, we see that incorporating demonstrations generally enhances performance for most tasks and models. Nevertheless, in many instances, particularly with GPT3.5, using just a single demonstration surprisingly leads to poorer performance. This could imply that a single demonstration might introduce a bias detrimental to performance (e.g., the LLM predicts the same label as the demonstration).

For tasks other than SA, we observe that a greater number of demonstrations tends to improve both in-domain and cross-domain performance. The influence on performance drops is less straightforward - it appears that increasing the number of demonstrations may either exacerbate the drop in performance or have no significant effect.

In conclusion, it is better to use a greater number of demonstrations, with a preference for those originating from the target domain.

\begin{figure*}[!htb]
    \centering
    \includegraphics[width=\textwidth]{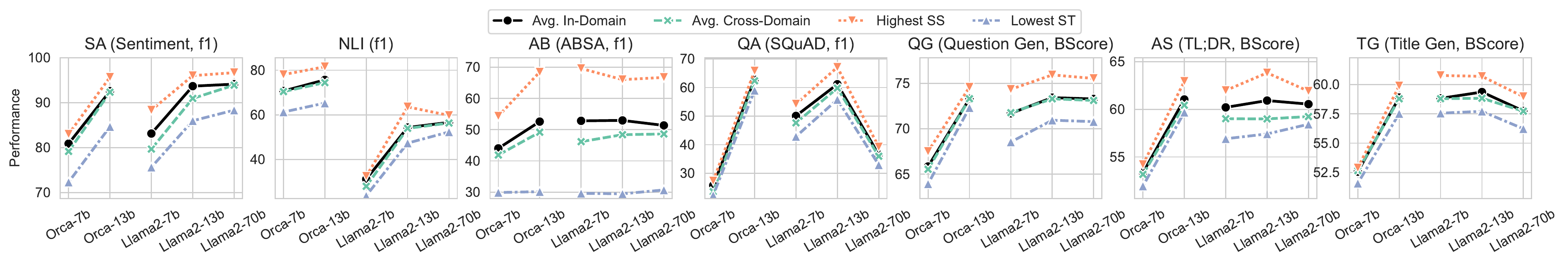}
    \caption{Few-shot (4 demonstrations) performance for the seven tasks of Llama2-family models with varying sizes. The plots present the F1 and BertScore scores of the average in-domain (black line) and cross-domain (green line) performance. In addition, the highest in-domain score (orange line) and the lowest cross-domain score (blue line) are displayed.
    Due to hardware constraints, all models were loaded with NF4 quantization, and computations were performed in 16-bit FP.}
    \label{fig:fs_performance}
\end{figure*}
\begin{figure*}[!htb]
    \centering
    \includegraphics[width=\textwidth]{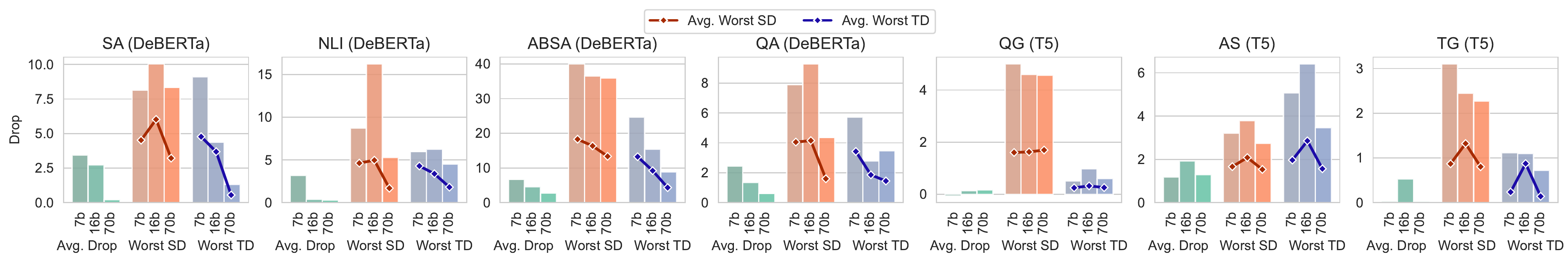}
    \caption{Few-shot (4 demonstrations) drops for the seven tasks of Llama2-family models with varying sizes. The plots present: The Average Drop (green bars); The Worst $\sd$ (orange bars); and the Worst $\td$ (blue bars). The lines on the bars present the Average Worst $\sd$ and $\td$, i.e., for each source domain we first find the worst drop and then take the average over all source domains.
    Due to hardware constraints, all models were loaded with NF4 quantization, and computations were performed in 16-bit FP.}
    \label{fig:fs_drops}
\end{figure*}

\subsection{Few-shot Model Size}
\label{sub:fs_model_size}

In this subsection, we explore the effect of the few-shot model size on DR. For this analysis, we experimented with LLMs from the Orca and Llama2 families. These families support 2 (Orca) and 3 (Llama2) of different sizes, all of which have undergone similar training and alignment procedures. 
Due to hardware constraints, we were unable to load the Llama2-70b model. Therefore, all Llama2 models were loaded with NF4 quantization, and computations were performed in 16-bit FP.

Although the results are inconclusive, since in some tasks (QA and TG) the performance of the 70b model sharply drops, we can still observe in Figure~\ref{fig:fs_performance} that increasing the model size generally improves the absolute in-domain and cross-domain performance. This behavior is not surprising and is similar to what is observed in fine-tuning setups. Regarding the drops presented in Figure~\ref{fig:fs_drops}, the trends can be mixed. Yet, it appears that both the average drops and the worst drops are decreasing as the size increases.

\begin{figure*}[!htb]
    \centering
    \includegraphics[width=\textwidth]{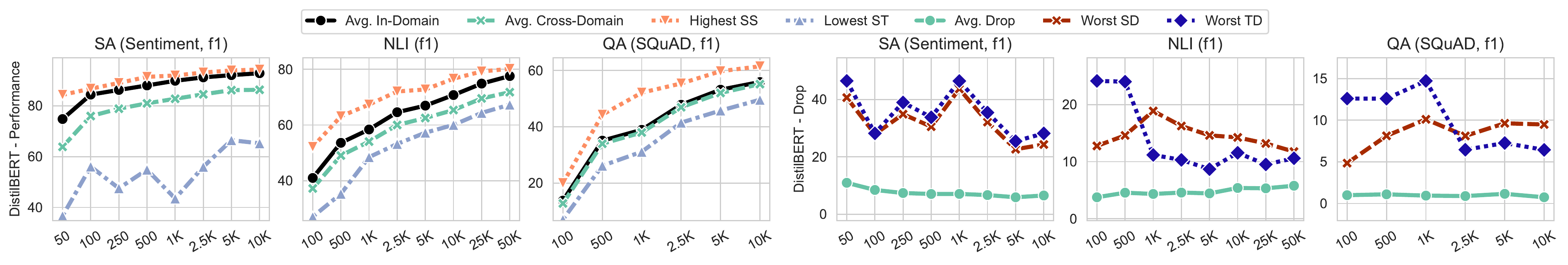}
    \includegraphics[width=\textwidth]{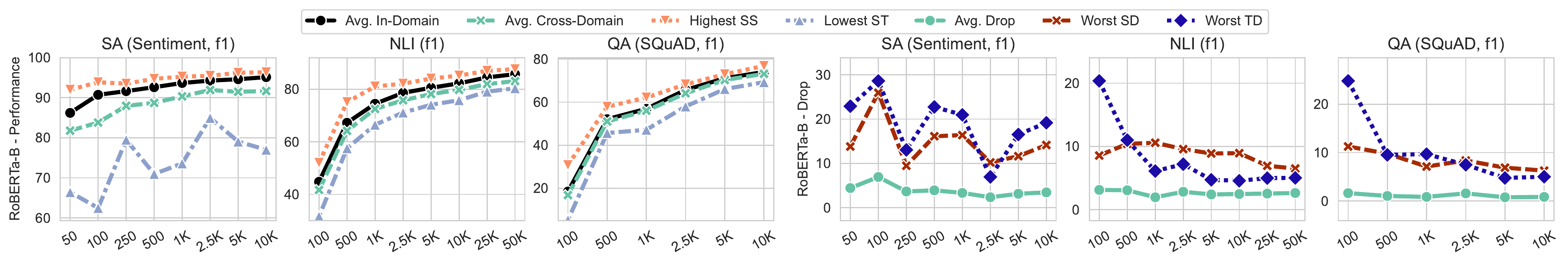}
    \includegraphics[width=\textwidth]{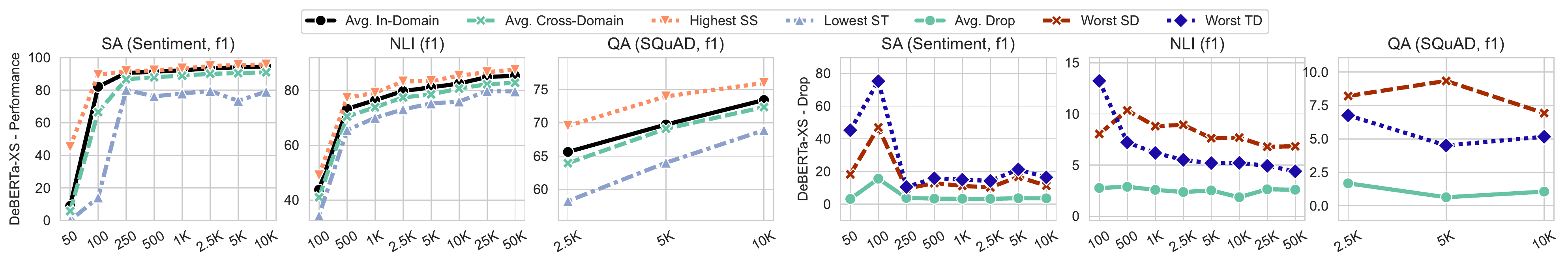}
    \includegraphics[width=\textwidth]{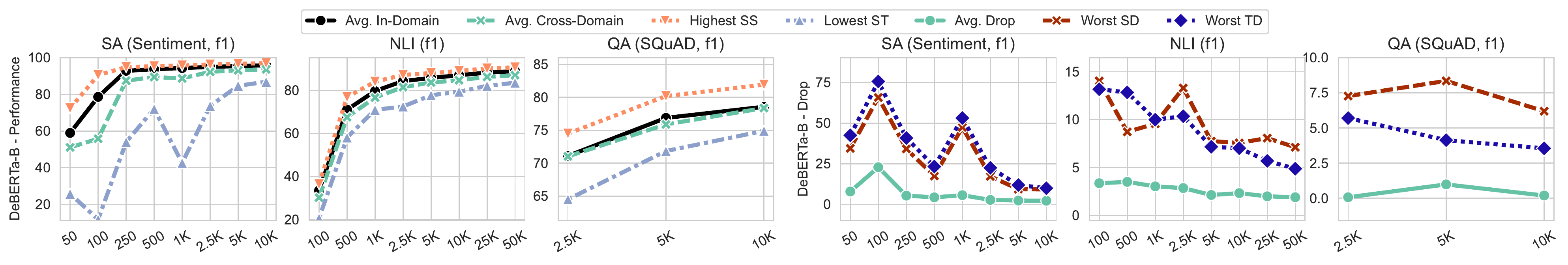}
    \caption{Classification performance and drops of DistilBERT (first row), RoBERTa-B (second row), DeBERTa-XS (third row) and DeBERTa-B (fourth row) as a function of the training dataset size of the source domain. In the leftmost three columns: The F1 scores of the average performance in-domain (black line); cross-domain (green line); The highest in-domain score (orange line); The lowest cross-domain score (blue line). In the rightmost three columns: The Average Drop (green line); The Worst $\sd$ (orange line); and the Worst $\td$ (blue line).}
    \label{fig:class_sizes}
\end{figure*}
\begin{figure*}[!htb]
    \centering
    \includegraphics[width=\textwidth]{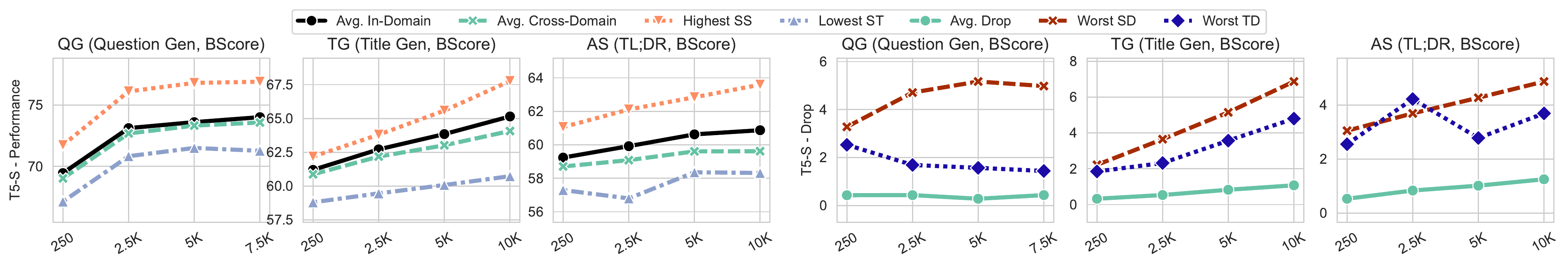}
    \includegraphics[width=\textwidth]{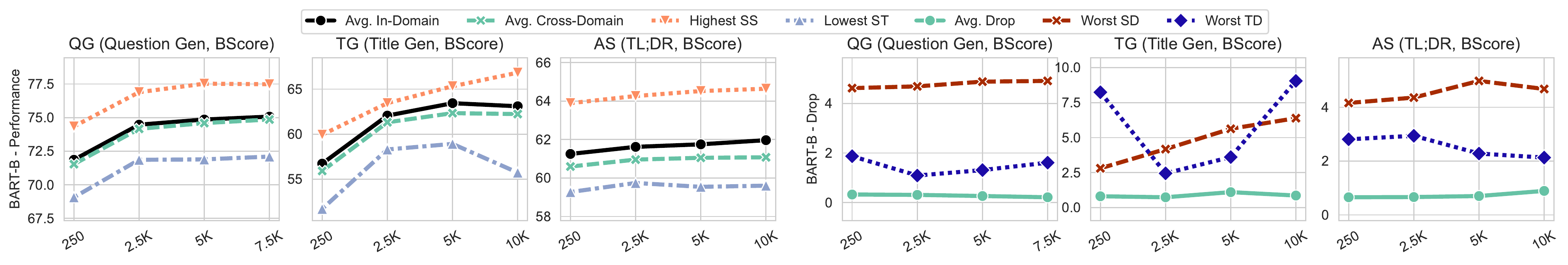}
    \includegraphics[width=\textwidth]{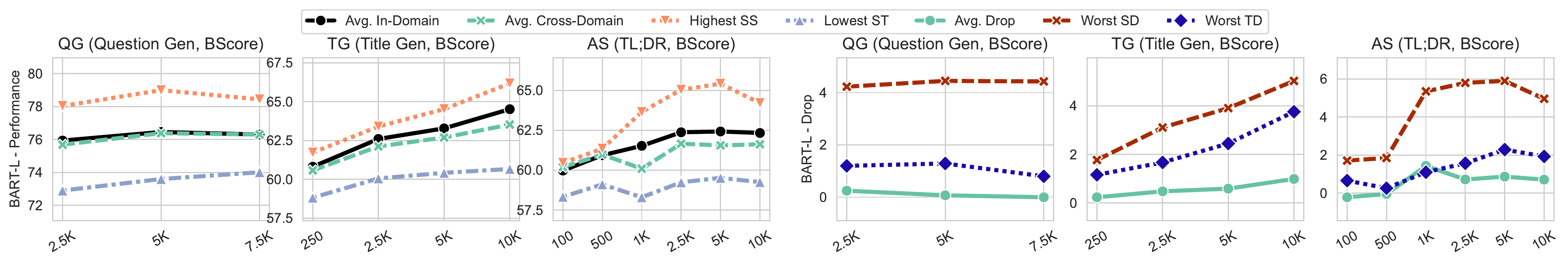}
    \caption{Generation performance and drops of T5-S (first row), BART-B (second row) and BART-L (third row) as a function of the training dataset size of the source domain. In the leftmost three columns: The BERTScores of the average performance in-domain (black line); cross-domain (green line); The highest in-domain score (orange line); The lowest cross-domain score (blue line). In the rightmost three columns: The Average Drop (green line); The Worst $\sd$ (orange line); and the Worst $\td$ (blue line).}
    \label{fig:gen_sizes}
\end{figure*}

\subsection{Dataset Size}
\label{sub:dataset_size}

Our next analysis aims to explore how the number of training samples from the source domain influences the domain robustness. Figures~\ref{fig:class_sizes} and ~\ref{fig:gen_sizes} depict the impact of the size of the source training dataset on the performance of models in classification and generation tasks, respectively.

As expected, an increase in the dataset size enhances performance in both in-domain and cross-domain. For classification tasks, while an increase in sample size tends to decrease the worst $\sd$ and $\td$, it does not affect the average drop. On the other hand, in generation tasks, the effect varies across different tasks. Interestingly, in the TG and AS tasks, we observe larger drops when increasing the number of samples.

\begin{figure*}[!htb]
    \centering
    \includegraphics[width=\textwidth]{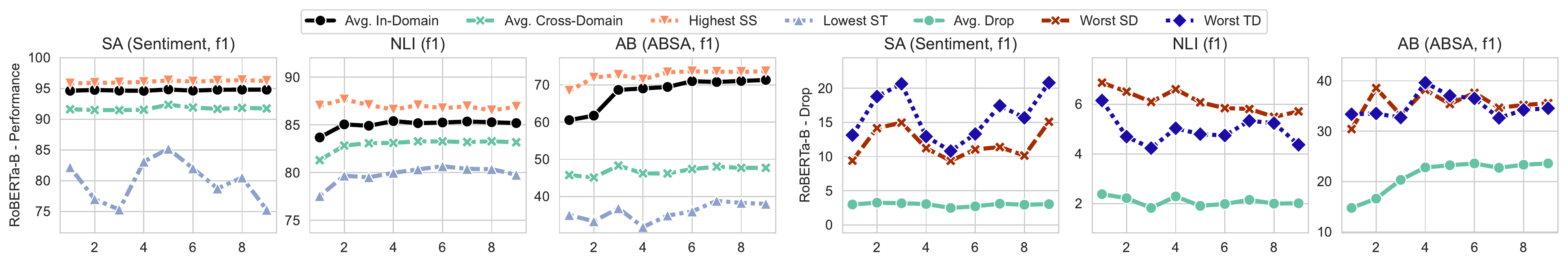}
    \includegraphics[width=\textwidth]{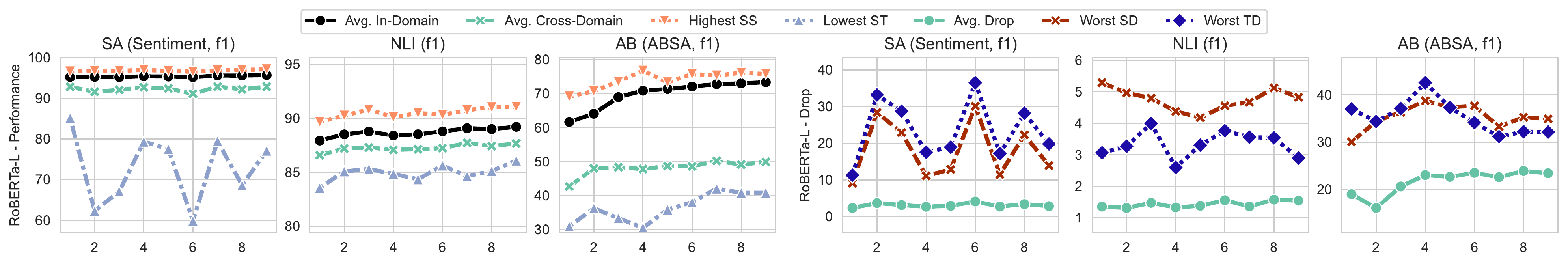}
    \includegraphics[width=\textwidth]{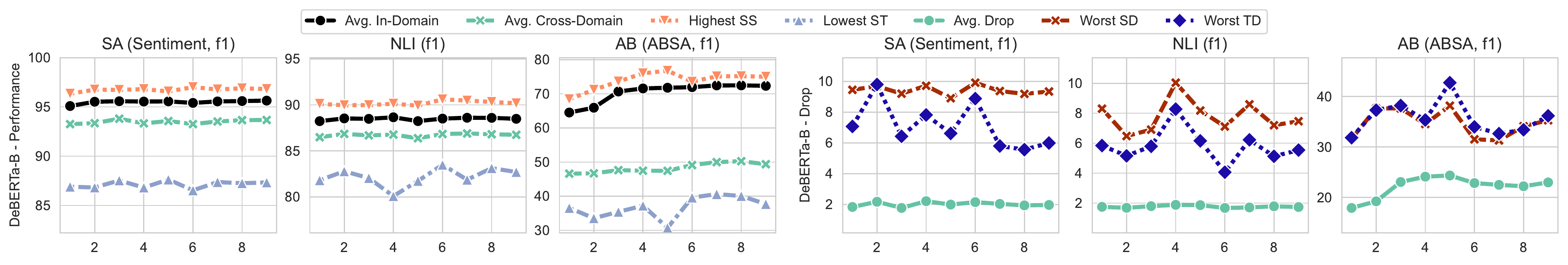}
    \includegraphics[width=\textwidth]{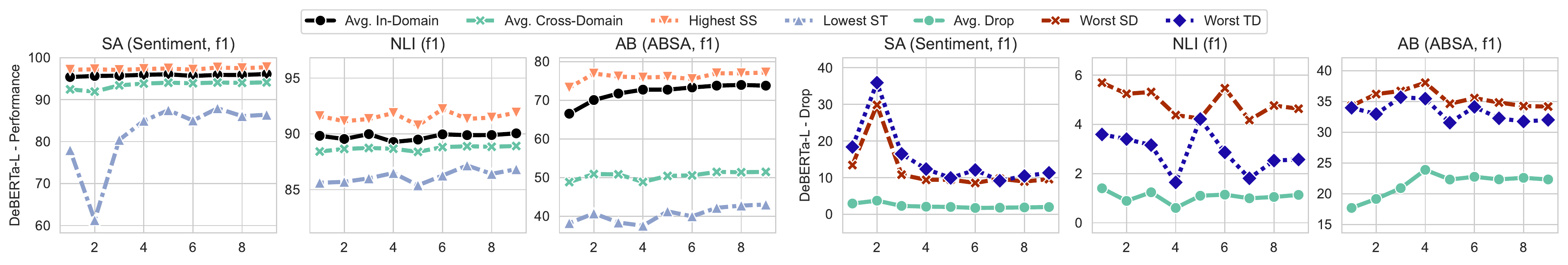}
    \caption{Performance and drops of RoBERTa-B (first row), RoBERT-L (second row), DeBERTa-B (third row) and DeBERTa-L (fourth row) as a function of the epoch. In the leftmost three columns: The F1 scores of the average performance in-domain (black line); cross-domain (green line); The highest in-domain score (orange line); The lowest cross-domain score (blue line). In the rightmost three columns: The Average Drop (green line); The Worst $\sd$ (orange line); and the Worst $\td$ (blue line).}
    \label{fig:epochs}
\end{figure*}

\subsection{Epochs and Model Selection}
\label{sub:epochs}

In the standard fine-tuning process, a model is trained until it no longer shows improvement on the validation set and the model selected for deployment is the one that attains the highest validation score. However, this approach does not guarantee optimal performance in the target domain, nor does it necessarily lead to the best model selection.  We therefore wish to measure how the in-domain and cross-domain performance evolve over the course of the fine-tuning procedure, across different epochs. 

As seen in Figure~\ref{fig:epochs}, in most cases, models appear to reach convergence in terms of average in-domain and cross-domain performance within a few epochs. Yet, it is noteworthy that the lowest cross-domain performance exhibits significant variability, undergoing substantial fluctuations during the training process. A similar pattern is observed in the performance drops.

These findings raise an interesting research question: Considering the significant variability of the cross-domain performance during the fine-tuning process, what is the optimal strategy for selecting a domain robust model? This question opens an interesting avenue for further research.

\subsection{Token Embeddings}
\label{sub:embeddings}

Every Transformer-based model employs an embedding matrix to transform tokens into continuous vectors. One strategy, known as `freezing' this matrix, involves not updating its weights during fine-tuning \citep{perl}. This tactic is motivated by the idea that, given the vocabulary differences across domains, maintaining the original embeddings might prevent the introduction of biases specific to the source domain. Consequently, this approach could potentially enhance the ability to generalize across different domains.  

The results, presented in Table \ref{tab:frozen}, indicate that freezing embeddings during fine-tuning does not harm the in-domain performance while increasing the cross-domain performance by approximately 0.5 points in SA and 0.2 points in NLI. Regarding the worst drops, in the SA task this approach remarkably improves the drops while in the NLI task, it slightly degrades them. These findings suggest that freezing embeddings could serve as a simple baseline for future research in domain adaptation.

\begin{table}[!htb]
\centering
\begin{adjustbox}{width=0.44\textwidth}
\begin{tabular}{l|ccccc}
\toprule
\textbf{Task} & {\footnotesize $\overline{\sss}$ } & {\footnotesize $\overline{\st}$ } & {\footnotesize $\drop$ } & {\footnotesize $W_{\sd}$ } & {\footnotesize $W_{\td}$ } \\
\midrule
  SA &           95.13 &              91.67 &       3.46 &     14.16 &     19.18 \\
  + FZ &           95.13 &              92.17 &       2.96 &     10.04 &     12.99 \\
\hline
 NLI &           85.76 &              83.13 &       2.63 &      6.51 &      5.03 \\
 + FZ &           85.60 &              83.33 &       2.27 &      7.04 &      5.18 \\
\bottomrule
\end{tabular}
\end{adjustbox}
\caption{Results of RoBERTa-B in the SA and NLI tasks under two scenarios: First, when the token embeddings matrix is trainable (SA and NLI), and second, when it is frozen (+ FZ). The columns are: $\overline{\sss}$ - Average In-domain Performance, $\overline{\st}$ - Average Cross-domain Performance, $\drop$ - Average Drop, $W_{\sd}$ - Worst $\sd$ and $W_{\td}$ - Worst $\td$.}
\label{tab:frozen}
\end{table}
\begin{table}[!htb]
\centering
\begin{adjustbox}{width=0.44\textwidth}
\begin{tabular}{l|l|ccccc}
\toprule
\textbf{Model} & \textbf{Task} & {\footnotesize $\overline{\sss}$ } & {\footnotesize $\overline{\st}$ } & {\footnotesize $\drop$ } & {\footnotesize $W_{\sd}$ } & {\footnotesize $W_{\td}$ } \\
\midrule
\multirowcell{2}{DistilBERT} &   QA &           55.89 &              55.15 &       0.75 &      9.49 &      6.44 \\
 &  + IB &           53.76 &              54.13 &      -0.37 &      7.51 &     17.62 \\
\hline
 \multirowcell{2}{RoBERTa-B} &   QA &           74.01 &              73.15 &       0.86 &      6.29 &      5.03 \\
 &  + IB &           70.66 &              69.96 &       0.71 &     15.67 &     24.42 \\
 \hline
 \multirowcell{2}{RoBERTa-L} &   QA &           82.01 &              81.72 &       0.29 &      6.01 &      2.53 \\
 &  + IB &           81.09 &              80.62 &       0.47 &     13.74 &     10.55 \\
 \hline
\multirowcell{2}{DeBERTa-XS} &   QA &           73.41 &              72.36 &       1.06 &      6.93 &      5.16 \\
 &  + IB &           70.50 &              70.01 &       0.48 &     17.86 &     21.51 \\
\hline
 \multirowcell{2}{DeBERTa-S} &   QA &           71.83 &              71.19 &       0.64 &      6.10 &      6.19 \\
 &  + IB &           66.10 &              67.00 &      -0.90 &     13.67 &     18.10 \\
 \hline
 \multirowcell{2}{DeBERTa-B} &   QA &           78.56 &              78.37 &       0.19 &      6.20 &      3.55 \\
 &  + IB &           78.57 &              78.21 &       0.36 &     14.95 &     10.00 \\
 \hline
 \multirowcell{2}{DeBERTa-L} &   QA &           74.54 &              74.10 &       0.44 &      6.29 &      2.72 \\
 &  + IB &           79.82 &              79.06 &       0.77 &     17.67 &     15.84 \\
\bottomrule
\end{tabular}
\end{adjustbox}
\caption{Results of fine-tuned models in the QA task under two scenarios: First, when all domains have an identical ratio of questions without answers (QA), and second, when the distribution of 'no answer' questions varies between domains (+ IB - imbalanced). The columns are: $\overline{\sss}$ - Average In-domain Performance, $\overline{\st}$ - Average Cross-domain Performance, $\drop$ - Average Drop, $W_{\sd}$ - Worst $\sd$ and $W_{\td}$ - Worst $\td$.}
\label{tab:imbalanced}
\end{table}

\subsection{Prior Shift}
\label{sub:imbalance}

When developing our benchmark, we decided to restrict it to several technical assumptions (described in \S\ref{sub:benchmark_assumptions}). These assumptions enable a precise and clear analysis in a ``controlled experiment'' manner. One of the assumptions is that the prior distribution $P(Y)$ remains relatively consistent across various domains. For classification tasks, every domain has the same class distribution. In the QA task, it translates to each domain having the same ratio of `no answer' examples (0.2). This subsection explores what happens when this assumption does not hold and a prior shift occurs. To this end, we reconstruct the QA dataset by resampling examples from each domain, reflecting their original `no answer' distribution. Accordingly, the ratio of `no answer' examples can vary between 0.05 and 0.4. 

In Table~\ref{tab:imbalanced}, we present the results of several encoder-only models trained on the balanced and imbalanced QA datasets. Our observations indicate that while the impact on the average is relatively low, the worst drops are much more prominent when the prior shift occurs. We analyzed the results and found a simple explanation for this.

The increased diversity across different domains leads to greater variability in absolute performance. For example, domains with a higher proportion of `no answer' questions, which are typically more challenging, tend to have a lower absolute in-domain performance (or lower cross-domain performance when shifting to those domains). This increased variability leads to more pronounced discrepancies between in-domain and cross-domain performance, resulting in larger drops. Although the average drop remains consistently low -- because sometimes the shift is to an easier domain, compensating drops when the shift is to a harder domain -- the worst drops are significantly more pronounced. This experiment effectively illustrates that as the shift becomes more prominent (affecting both X and Y variables), there is a notable increase in performance variability across domains, leading to more substantial drops in some cases.

\subsection{Scenarios Statistical Validation}
\label{sub:scenarios_stats}

In \S\ref{sub:scenarios} we introduce four possible scenarios of domain shift: Classic, Unobserved, Observed, and No Challenge. Each scenario is determined by the sign of the $\sd$ and the $\td$ of a single domain shift. We present the proportion of each scenario in Figure~\ref{fig:scenarios}, taking into account the results of all domain shifts and all participating models. For all few-shot models, we use 4-shots. Since we conducted experiments with more shots in \S\ref{sub:number_demonstrations}, we also include results 8-shots for GPT3.5, Neural, and Mistral, and 16-shots for GPT3.5. 

We next validate whether the domain shift has a statistically valid effect on the model performance. Consider that if there is no effect, we would expect the order of $(\sss, \ttt, \st)$ to be distributed uniformly. There are six possible sequences, where two belong to the Classic scenario ($\st < \sss < \ttt$ or $\st < \ttt < \sss$), two belong to the No Challenge scenario ($\sss < \ttt < \st$ or $\ttt < \sss < \st$), one belongs to the Observed scenario ($\ttt < \st < \sss $), and one to the Unobserved scenario ($\sss < \st < \ttt $). Under the assumption of uniform distribution, each sequence would have a probability of $\sfrac{1}{6}$. 

We conduct a Chi-square test with a significance threshold of 0.05, applying a Bonferroni correction for multiple comparisons (14 tests in total, adjusting the significance level to 0.0036). The test results show that all P-values are below 0.001, except for the QA task in few-shot models, which is at 0.004. These findings confirm that the effect of domain shift on model performance is statistically significant. Notably, the results highlight that the demonstration domain used in few-shot models influences the cross-domain performance.

        
    
\begin{table*}[!htb]
\begin{tabular}{|p{\textwidth}<{\centering}|}         
\hline
               
\rowcolor{gray!60}               
\textbf{Motivation} \\               
\footnotesize
\begin{tabular}{p{0.23\textwidth}<{\centering} p{0.23\textwidth}<{\centering} p{0.23\textwidth}<{\centering} p{0.23\textwidth}<{\centering}}                        
\textit{Practical} & \textit{Cognitive} & \textit{Intrinsic} & \textit{Fairness}\\
$\square$		
& 		
& 		
& 		
\\
\end{tabular}\\
               
\rowcolor{gray!60}               
\textbf{Generalisation type} \\               
\footnotesize
\begin{tabular}{p{0.14\textwidth}<{\centering} p{0.14\textwidth}<{\centering} p{0.15\textwidth}<{\centering} p{0.15\textwidth}<{\centering} p{0.15\textwidth}<{\centering} p{0.15\textwidth}<{\centering}}                   
\textit{Compositional} & \textit{Structural} & \textit{Cross Task} & \textit{Cross Language} & \textit{Cross Domain} & \textit{Robustness}\\
& 		
& 		
& 		
& $\square$		
& $\square$		
\\
\end{tabular}\\
             
\rowcolor{gray!60}             
\textbf{Shift type} \\             
\footnotesize
\begin{tabular}{p{0.23\textwidth}<{\centering} p{0.23\textwidth}<{\centering} p{0.23\textwidth}<{\centering} p{0.23\textwidth}<{\centering}}                        
\textit{Covariate} & \textit{Label} & \textit{Full} & \textit{Assumed}\\  
$\square$		
& 		
& 		
& 		
\\
\end{tabular}\\
             
\rowcolor{gray!60}             
\textbf{Shift source} \\             
\footnotesize
\begin{tabular}{p{0.23\textwidth}<{\centering} p{0.23\textwidth}<{\centering} p{0.23\textwidth}<{\centering} p{0.23\textwidth}<{\centering}}                          
\textit{Naturally occuring} & \textit{Partitioned natural} & \textit{Generated shift} & \textit{Fully generated}\\
$\square$		
& 		
& 		
& 		
\\
\end{tabular}\\
             
\rowcolor{gray!60}             
\textbf{Shift locus}\\             
\footnotesize
\begin{tabular}{p{0.23\textwidth}<{\centering} p{0.23\textwidth}<{\centering} p{0.23\textwidth}<{\centering} p{0.23\textwidth}<{\centering}}                         
\textit{Train--test} & \textit{Finetune train--test} & \textit{Pretrain--train} & \textit{Pretrain--test}\\
$\square$		
& $\square$		
& 		
& $\square$		
\\
\end{tabular}\\
\hline
\end{tabular}
\caption{Categorization of our study according to the GenBench taxonomy \citep{da_hu23}.}
\end{table*}
\section{The Domain Robustness Benchmark: Technical  Details}
\label{sec:benchmark_technical}

\subsection{Preprocessing}
\label{sub:preprocessing}

\medskip\noindent\textbf{Sentiment Analysis (SA)} We removed links from texts since they were tokenized to dozens of tokens and significantly increased the input length.

\medskip\noindent\textbf{Question Answering (QA)} We split the documents of each category (and their corresponding questions) into train, development, and test sets. 

\medskip\noindent\textbf{Question Generation (QG)} The input is a concatenation of the document and the answer, separated by the ``answer:'' token.

\medskip\noindent\textbf{Abstractive Summarization (AS)} 
Since the summaries of the Webis-TLDR-17 dataset were automatically extracted and not verified, they may be of low quality. After manually examining dozens of them, we decided to use only summaries that have 15-60 words, and at least 75\% of them appear in the post. 

\medskip\noindent\textbf{Title Generation (TG)} After manually examining examples, we found many reviewers misused the title option: They started writing a long review in the title and continued it in the body box. We therefore decided to use only titles that have 5-20 words, and at least 75\% of them appear in the grounding review.

\subsection{Technical Domain Shift Assumptions}
\label{sub:benchmark_assumptions}

As discussed in \S\ref{sec:methods}, a domain can be characterized by various attributes such as topic, style, syntax, and medium. When one of these attributes changes, the joint distribution $P(X,Y)$ changes, and a domain shift occurs.
In developing our benchmark, we grounded it in technical assumptions aimed at facilitating a controlled experimental analysis, as detailed \S\ref{sub:benchmark_assumptions}. One of these assumptions is to focus on natural topic shifts (although other factors are likely to change as well, such as the style and syntax). This contrasts with other studies that explore synthetic shifts, such as adversarial attacks, challenge sets, or transitions to datasets from different data-generating processes (e.g., having other annotation guidelines). 

Our rationale was to isolate and control a single variable and facilitate a ``controlled experiment'' approach, allowing for a precise and clear analysis and characterization of the DR challenge. In line with this objective, we have established the following technical assumptions: 
\begin{enumerate}
    \item Our benchmark focuses on natural topic shift, e.g., training an NLP model on book reviews and applying it to kitchen product reviews. 
    In contrast to many other works \citep{transformers_improve_robustness, robustness_qa, wilds, revisiting}, our natural topic shift allows us to avoid complexities that arise when the shift is a byproduct of constructing a challenge set or transitioning to another dataset that was constructed by a different data generating process (e.g., different annotation guidelines). 
    \item Each task consists of several domains, facilitating a more comprehensive and accurate estimation of average performance and performance degradation.
    \item For each task, all the domains have the same number of training examples, enabling its use as a source and as a target domain. Moreover, it helps mitigate (non-DR) biases that may arise when transitioning from a domain with sufficient training data to a domain with scarce labeled data.
    \item We try to reduce the effect of the prior shift, i.e., changes in $P(Y)$: For classification tasks, we create balanced datasets (for QA, same ratio of `no answer'), while for generation tasks, we sample examples with similar output length distributions. In Appendix \S\ref{sub:imbalance}, we discuss experiments exploring changes in $P(Y)$ upon a domain shift. We found that this variation leads to increased performance variability across domains, resulting in larger worst drops but minimally impacting the average drop (because shifts to easier domains compensate for shifts to harder domains). 
\end{enumerate}

While our assumptions simplify the domain shift, we argue that if the DR challenge exists under these assumptions (and it does), then it will definitely exist more severely when our assumptions are violated and a complex shift occurs. 
Researchers who wish to focus on a specific type of prior shift (e.g., unbalanced domains) can easily use our publicly available benchmark to construct more challenging setups.

\subsection{Domains for Few-shot Experiments} 
\label{sub:fs_domains}
As mentioned in Section \ref{sec:experimental_setup}, due to the high costs associated with API calls, we limit our presentation of few-shot results to only three domains for each task, rather than encompassing all five or six domains. In addition, we randomly sample 200 test examples for each target domain. This cost constraint arises from the quadratic increase in the number of experiments relative to the number of domains (for instance, six domains lead to 36 domain-shift setups, whereas three domains result in just 9). Additionally, the extended input length, a consequence of augmenting it with multiple demonstrations, also contributes to this decision. For a fair comparison, we present results for the same three domains for both few-shot and fine-tuned models in Table~\ref{tab:main_fs_results}. The specific domains we focus on are:
\begin{itemize}
    \item SA - Airline, Beauty, Books.
    \item NLI - Fiction, Telephone, Traval.
    \item AB - Device, Laptops, MAMs.
    \item QA - History, Science, Society. 
    \item QG - Geography, History, Science. 
    \item AS - Fitness, LoL, Relationships.
    \item TG - Beauty, Books, DVDs.
\end{itemize}

\section{Implementation Details}
\label{sec:implementation}

Our experiments are conducted in the PyTorch and HuggingFace frameworks and optimize the fine-tuning models with the AdamW optimizer. An exception is the OpenAI's models, which were run via their paid API service and their results are correct as of January 2023. The data, results and code are provided in the project repository. 

\medskip\noindent\textbf{Hyperparameter Tuning}
For each model and source domain, we initially conduct hyperparameter tuning, selecting the optimal set based on the source domain's validation set. Subsequently, we evaluate the model across all target domains. In the hyperparameter tuning phase for classification models, we experiment with the following learning rates: [1e-5, 5e-5, 1e-4] and batch sizes: [4, 8, 16, 64] and 10 epochs. For generation models, we explore learning rates of [1e-3, 5e-4, 1e-4, 5e-5, 1e-5], use a batch size of 64 and 15 epochs.

\medskip\noindent\textbf{Instructions and Demonstrations}
For each test example from a target domain, the LLM input includes a system prompt detailing the task instruction, and a user prompt presenting the example. In few-shot setups, we augment this with additional demonstrations (input and target) from the source domain. This involves adding extra user-assistant turns: the user turn shows the demonstration input, and the assistant turns present the demonstration target (label). We randomly select demonstrations from the source domain's training set for each test. In classification tasks, for $N>1$-shots the prompt includes demonstrations of all labels. Task instructions and prompt examples are in Appendix~\ref{sub:prompts}. In addition, to not exceed the maximum input length of several models, we truncate the maximum length of each demonstration to 256 tokens (but no truncation was applied to the test example). Please see L2 in \S\ref{sec:limitations} for other prompting attempts.

The classification results of few-shot LLMs are based on ``long-form generation''. Notice that we mentioned the labels in the prompt and asked the LLM to respond only with them (see examples \S\ref{sub:prompts}). The LLMs we used in our study underwent SFT with instructions and, therefore, almost always followed our instructions and responded with a label (we also used temperature=0.0). When they did not--such as when they began generating an explanation before or after stating the label--we extracted the first mentioned label (lowercase). We found labels 100\% of the time (except for CodeLlama-70b).



\subsection{Prompts}
\label{sub:prompts}

\begin{tcolorbox}[enhanced,breakable,colback=red!5!white,colframe=red!75!black,title=\textbf{Prompt for SA (Sentiment Analysis)}]
\texttt{SYSTEM}\\You will be provided with a review and asked to classify its sentiment.\\
You can only response "negative" or "positive".\\\\
\texttt{USER}\\Review:\\\texttt{[text]}\\
\end{tcolorbox}

\begin{tcolorbox}[enhanced,breakable,colback=orange!5!white,colframe=orange!75!black,title=\textbf{Prompt for NLI (Multi-NLI)}]
\texttt{SYSTEM}\\You will be provided with a premise and a hypothesis and asked to classify their relationship.\\
You can only response "entailment", "neutral" or "contradiction".\\\\
\texttt{USER}\\Premise:\\\texttt{[premise]}\\\\Hypothesis:\\\texttt{[hypothesis]}\\
\end{tcolorbox}

\begin{tcolorbox}[enhanced,breakable,colback=violet!5!white,colframe=violet!75!black,title=\textbf{Prompt for AB (ABSA)}]
\texttt{SYSTEM}\\You will be provided with a sequence of words and asked to extract the aspect and the polarity of each word.\\
You can only response with a sequence of tags corresponding to each word. The tags are: "O", "T-POS", "T-NEG", "T-NEU", where "O" indicates a non aspect word. For example, the answer of: "The good boy", is: "O O T-POS".\\\\
\texttt{USER}\\Text:\\\texttt{[text]}\\
\end{tcolorbox}

\begin{tcolorbox}[enhanced,breakable,colback=magenta!5!white,colframe=magenta!75!black,title=\textbf{Prompt for QA (SQuAD v2)}]
\texttt{SYSTEM}\\You will be provided with a context and a question and asked to extract the answer from the context.\\
You can only response with a copied span of text from the context. If there is no answer, response: "No answer". \\\\
\texttt{USER}\\Context:\\\texttt{[context]}\\\\Question:\\\texttt{[question]}\\
\end{tcolorbox}

\begin{tcolorbox}[enhanced,breakable,colback=lime!5!white,colframe=lime!75!black,title=\textbf{Prompt for QG (Question Generation)}]
\texttt{SYSTEM}\\You will be provided with a context and an answer, and asked to generate a question that would lead to the answer.\\
You can only response with the question. \\\\
\texttt{USER}\\Context:\\\texttt{[context]}\\\\Answer:\\\texttt{[answer]}\\
\end{tcolorbox}

\begin{tcolorbox}[enhanced,breakable,colback=teal!5!white,colframe=teal!75!black,title=\textbf{Prompt for AS (TL;DR Abstractive Summarization)}]
\texttt{SYSTEM}\\You will be provided with a reddit post and asked to generate a short TL;DR summary of the post that the Redditor might have written at the end of the post.\\
You can only response with the summary. \\\\
\texttt{USER}\\Post:\\\texttt{[text]}\\
\end{tcolorbox}

\begin{tcolorbox}[enhanced,breakable,colback=cyan!5!white,colframe=cyan!75!black,title=\textbf{Prompt for TG (Title Generation)}]
\texttt{SYSTEM}\\You will be provided with a product review and asked to generate a title that the reviewer might have given to the review.\\
You can only response with the title. \\\\
\texttt{USER}\\Review:\\\texttt{[text]}\\
\end{tcolorbox}

\begin{tcolorbox}[enhanced,breakable,colback=brown!5!white,colframe=brown!75!black,title=\textbf{Example of 2-shot SA prompt}]
\texttt{SYSTEM}\\You will be provided with a review and asked to classify its sentiment.\\
You can only response "negative" or "positive".\\\\
\texttt{USER}\\Review:\\\texttt{[text1]}\\\\
\texttt{ASSISTANT}\\negative\\\\
\texttt{USER}\\Review:\\\texttt{[text2]}\\\\
\texttt{ASSISTANT}\\positive\\\\
\texttt{USER}\\Review:\\\texttt{[text]}\\
\end{tcolorbox}



\clearpage

\end{document}